%% file: arxiv.tex
\PassOptionsToPackage{table}{xcolor}
\documentclass{article} 
\usepackage{arxiv,times}

\input{math_commands.tex}

\colorlet{revisionblue}{black}
\usepackage{amssymb}
\usepackage{booktabs}
\usepackage{array}
\usepackage{tabularx}
\usepackage{graphicx}
\usepackage{adjustbox}
\usepackage{float}
\usepackage{microtype}
\usepackage{multirow}
\usepackage{enumitem} \usepackage{comment}
\usepackage{tikz}
\usetikzlibrary{arrows.meta,backgrounds,fit,positioning}
\usepackage{url}
\usepackage{hyperref}
\hypersetup{
    colorlinks=true,
    linkcolor=black,
    citecolor=black,
    urlcolor=black,
    pdftitle={TableLoop: Tabularizing Multimodal Data via Rubric Evolution},
pdfauthor={Yutong Bi, Carson Dudley, Ege Onur Taga, Xiaofeng Liu, Samet Oymak}
}

\newcommand{\topthree}{top-three ensemble}

\newcommand{\better}{\cellcolor{green!12}}

\newcommand{\metricup}{$\uparrow$}
\newcommand{\metricdown}{$\downarrow$}
\newcolumntype{Y}{>{\centering\arraybackslash}p{0.78cm}}

\title{%
  \parbox{\textwidth}{%
    \rule{\linewidth}{2.5pt}

    \vspace{0.12cm}

    \begin{minipage}[c]{0.16\textwidth}
      \centering
      \includegraphics[width=0.90\linewidth]{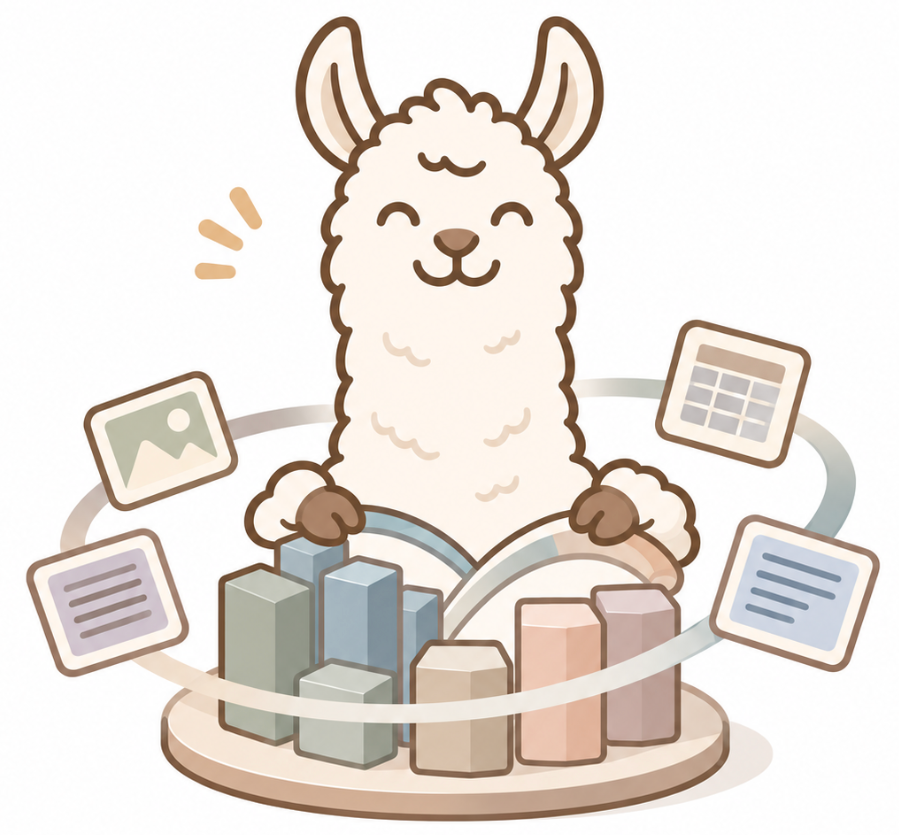}
    \end{minipage}%
    \hfill
    \begin{minipage}[c]{0.81\textwidth}
    \raggedright
    \textbf{TableLoop: Tabularizing Multimodal Data\\
    \hphantom{TableLoop: \ }via Rubric Evolution}
\end{minipage}

    \vspace{0.12cm}

    \rule{\linewidth}{1pt}
  }%
}

\author{
Yutong Bi$^{1}$,
Carson Dudley$^{1}$,
Ege Onur Taga$^{1}$,
Xiaofeng Liu$^{2,1}$,
Samet Oymak$^{1}$ \\[12pt]
$^{1}$University of Michigan \\
$^{2}$University of Massachusetts Boston \\[8pt]
\texttt{\{yutongb, cdud, egetaga, oymak\}@umich.edu} \\
\texttt{xiaofeng.liu@umb.edu}
}
\begin{document}

\maketitle

\begin{abstract}
Many prediction tasks depend on information distributed across text, images, and structured measurements. We introduce TableLoop, a flexible framework that connects language and vision-language models with tabular predictors through learned natural-language rubrics, closing the loop between what a model measures and how useful those measurements are for downstream prediction. A frozen scoring model applies each rubric to produce explicit numeric features, while a tabular predictor's validation performance and errors guide iterative revisions to the rubric{\color{revisionblue}'s criteria and scoring instructions}. The resulting interpretable features are combined with structured measurements for final prediction. TableLoop is modular: it requires no language-model tuning and supports interchangeable scoring models, tabular predictors, and evolution methods. Across nine tasks spanning clinical risk, persuasion outcomes, essay quality, peer-review decisions, skin-lesion malignancy, and fungal species identity, TableLoop outperforms both structured-data-only prediction and calibrated {\color{revisionblue}zero-shot} prediction by a larger language or vision-language model in 26 of 27 evaluated configurations {\color{revisionblue}across three evolution methods}. {\color{revisionblue}TableLoop also outperforms frozen-embedding baselines on 6 of 7 tasks and prompt-optimized direct predictors on all three tested tasks, while comparisons with fine-tuned models show task-dependent advantages. Ablations further demonstrate the value of iterative rubric refinement for downstream tabular prediction.} Notably, TableLoop using a smaller open-weight scorer (Qwen3-32B/Qwen3-VL-32B) can outperform direct prediction with a frontier model (Claude Sonnet 4.5), showing that a smaller model can support stronger predictions when used to extract task-specific features rather than predict the target directly. Beyond prediction, TableLoop produces reusable, interpretable feature tables grounded in real observations, offering a potential source of training and adaptation data for tabular foundation models and an inspectable interface for multimodal-to-table learning.
\end{abstract}

\section{Introduction}

Many prediction problems require combining unstructured data, such as natural language and images, with structured measurements \citep{shi2021benchmarking,zhang2020combining,tang2024automm}. For example, predicting a patient's medical outcome may require interpreting doctors' notes alongside vital signs and laboratory results \citep{wang2025snow}. These sources can provide complementary evidence, but using them together requires a model to both interpret unstructured inputs and learn how they relate to the recorded measurements. Language and vision language models offer a way to extract information from text and images, while tabular predictors learn relationships among explicit features. Connecting these capabilities offers a practical route to prediction from heterogeneous data.

One approach is to convert unstructured inputs into numeric features that a tabular predictor can combine with existing measurements \citep{bisercic2023temed,balek2025features}. Natural-language rubrics provide an explicit interface for this conversion: a rubric defines properties to assess and instructions for scoring them, and a language or vision-language model assigns a score to each property for every input \citep{hashemi2024llmrubric,lam2024lloom}. This makes the extracted information interpretable: researchers can inspect both what each feature is intended to measure and the scores assigned to individual inputs. The tabular representation also makes it straightforward to combine these features with metadata or add further measurements and features as they become available. However, a plausible rubric may not be the most useful one for prediction. Its criteria may be difficult to score consistently, redundant, or weakly related to the outcome of interest. Feature selection can identify useful features within an existing rubric \citep{guyon2003selection}, but it cannot introduce properties that the rubric does not measure or improve how they are defined and scored. Prior work has explored learning concepts and natural-language features from labeled examples and prediction errors \citep{ludan2024tbm,zhong2024nlparam,feng2025bcllm,jojic2025genz}, including approaches that construct rubric-based representations, optimize feature-generation instructions, or refine grading rubrics \citep{demirel2026representations,cosma2026feature,chu2026caro}. Our goal is to optimize a complete rubric—its criteria and scoring instructions—using feedback from a predictor trained on the resulting scores.

We introduce \textbf{TableLoop}, a framework for combining structured and unstructured data through rubrics learned from predictive feedback (Figure \ref{fig:tableloop}). A {\color{revisionblue}frozen} language or vision language model applies each candidate rubric to produce numeric features, a tabular predictor learns from these features, and its performance and errors on validation data guide a rubric-evolution model in revising the scoring criteria and instructions. TableLoop supports interchangeable scoring models, tabular predictors, and rubric-evolution methods \citep{khattab2024dspy,guo2024evoprompt,agrawal2025gepa}. We also introduce a novel method for rubric optimization: successive halving with complementary merge (SH-merge), which repeatedly combines candidate rubrics and evaluates the resulting smaller population on progressively more data.

The output of TableLoop is not just a predictor. Each learned rubric defines an explicit procedure for converting unstructured observations into compatible tabular features, and the resulting tables can be inspected, stored, and reused with different predictive models. Because rubric criteria are expressed in natural language, the framework can be applied across domains without requiring a manually specified feature set for each task. This also creates an opportunity to construct tabular datasets from existing labeled text and image collections, potentially broadening the data available for training and adapting tabular foundation models.

We evaluate TableLoop on nine prediction tasks spanning clinical risk, persuasion outcomes, essay quality, peer-review decisions, skin-lesion malignancy, and fungal species identity. With TabPFN as the tabular predictor \citep{hollmann2025tabpfn}, TableLoop outperforms both a predictor using structured measurements alone and a larger language or vision language model making {\color{revisionblue} calibrated zero-shot} predictions directly from all available inputs {\color{revisionblue} in 26 of 27 configurations across the nine tasks and three evolution methods} (Figure \ref{fig:main-results}). {\color{revisionblue}TableLoop also outperforms frozen-embedding baselines on 6 of 7 tasks and is competitive with fine-tuned multimodal models, with particularly strong gains on persuasion and paper acceptance prediction. Comparisons with prompt-optimized direct predictors and rubrics generated without iterative refinement further demonstrate the value of combining rubric evolution with downstream tabular prediction. }These results show that a smaller language model coupled with learned rubrics and a tabular predictor can outperform prediction by a larger model while retaining explicit intermediate features for inspection, reuse, and interpretability.

Our contributions are threefold:
\begin{itemize}
\item \textbf{A flexible framework for combining structured and unstructured data.} TableLoop connects language and vision language models with tabular predictors through reusable, inspectable feature tables and supports interchangeable models and rubric-evolution methods.
\item \textbf{Learning what to measure through feedback.} We optimize rubric criteria and scoring instructions using validation performance. We also introduce SH-merge, which combines candidate rubrics while progressively increasing their evaluation budgets.
\item \textbf{Evaluation across tasks and design choices.} Across nine tasks we evaluate, TableLoop exceeds both metadata-only and direct-prediction baselines in each task. Ablations examine rubric initialization, evolution method, ensemble selection, structured-data integration, and predictor choice. {\color{revisionblue}Additional comparisons examine the contributions of iterative rubric refinement and tabularization. We also present rubric examples to illustrate the learned features and report runtime measurements for computational cost.}
\end{itemize}

\section{Related Work}

\paragraph{Combining structured and unstructured data.}
Prior work combines text and structured measurements through multimodal models or by converting structured records into text for language models \citep{shi2021benchmarking,tang2024automm,hegselmann2023tabllm}. {\color{revisionblue}Frozen text and image embeddings provide representations that can be combined with structured measurements~\citep{zhang2025qwen3, li2026qwen3}, while multimodal frameworks such as AutoMM fine tune encoders for downstream prediction~\citep{tang2024automm}.} An alternative is to convert unstructured inputs into explicit features for a tabular predictor. Language models can generate features by transforming existing columns \citep{hollmann2023caafe,han2024featllm,abhyankar2026llmfe} or extracting attributes from text \citep{bisercic2023temed,balek2025features,wang2025snow}, while tabular foundation models provide powerful predictors for the resulting tables \citep{hollmann2023tabpfn,hollmann2025tabpfn,tabfm2026}. Rubrics define these features through named criteria and scoring instructions, supporting inspection and concept exploration \citep{lam2024lloom,koh2020concept}. LLM-Rubric learns to combine scores from a fixed rubric \citep{hashemi2024llmrubric}, while other work constructs a rubric from labeled examples \citep{demirel2026representations}. TableLoop uses predictive feedback to revise the rubric itself, then combines its scores with structured measurements.

\paragraph{Learning features from predictive feedback.}
Several methods use labeled examples and prediction errors to learn concepts or natural-language features \citep{ludan2024tbm,zhong2024nlparam,feng2025bcllm,jojic2025genz,zheng2026structcbm}. TableLoop builds on this direction by optimizing a complete rubric, allowing its criteria and scoring instructions to change together. Text Bottleneck Models incrementally propose and refine concepts using misclassified examples, retaining those that improve a linear predictor \citep{ludan2024tbm}. FEST builds a bank of semantic and executable features from contrasts between positive and negative examples and uses decision-tree importance to guide which features remain in the bank \citep{khurana2026fest}. \citet{cosma2026feature} take a different approach, optimizing a prompt that tells a language model how to generate features.
 TableLoop instead optimizes the rubric that defines the features and how to score them. We compare these optimization targets in an appendix ablation, where direct rubric evolution performs better on both evaluated tasks (Appendix~\ref{sec:instruction-evolution}).
 Other work revises grading rubrics to improve an LLM's assigned grades \citep{chu2026caro}, whereas TableLoop optimizes rubric scores as inputs to a separate predictor. Combining learned concepts with structured information also has precedent: BC-LLM combines note-derived concepts with a tabular model's risk score \citep{feng2025bcllm}. TableLoop supplies the original structured measurements alongside rubric scores to the final predictor.

\paragraph{Optimizing rubrics efficiently.}
Prompt-optimization methods provide tools for proposing and revising instructions using performance feedback \citep{yang2024opro,guo2024evoprompt,fernando2024promptbreeder,khattab2024dspy,agrawal2025gepa}. {\color{revisionblue}Whereas direct prediction prompt optimization adapts instructions for predicting the target, TableLoop adapts the criteria and scoring instructions that define features for a separate tabular predictor.} TableLoop applies these methods to rubric search, where evaluating each candidate requires scoring observations and fitting a tabular predictor. Our SH-merge method draws on successive halving \citep{jamieson2016nonstochastic}: the candidate population shrinks across rounds while each candidate is evaluated on more data. Before reducing the population, it merges higher- and lower-ranked rubrics, allowing criteria from weaker candidates to contribute to subsequent rounds.

\section{Methods}

\begin{figure*}[t]
\centering
\includegraphics[width=\textwidth]{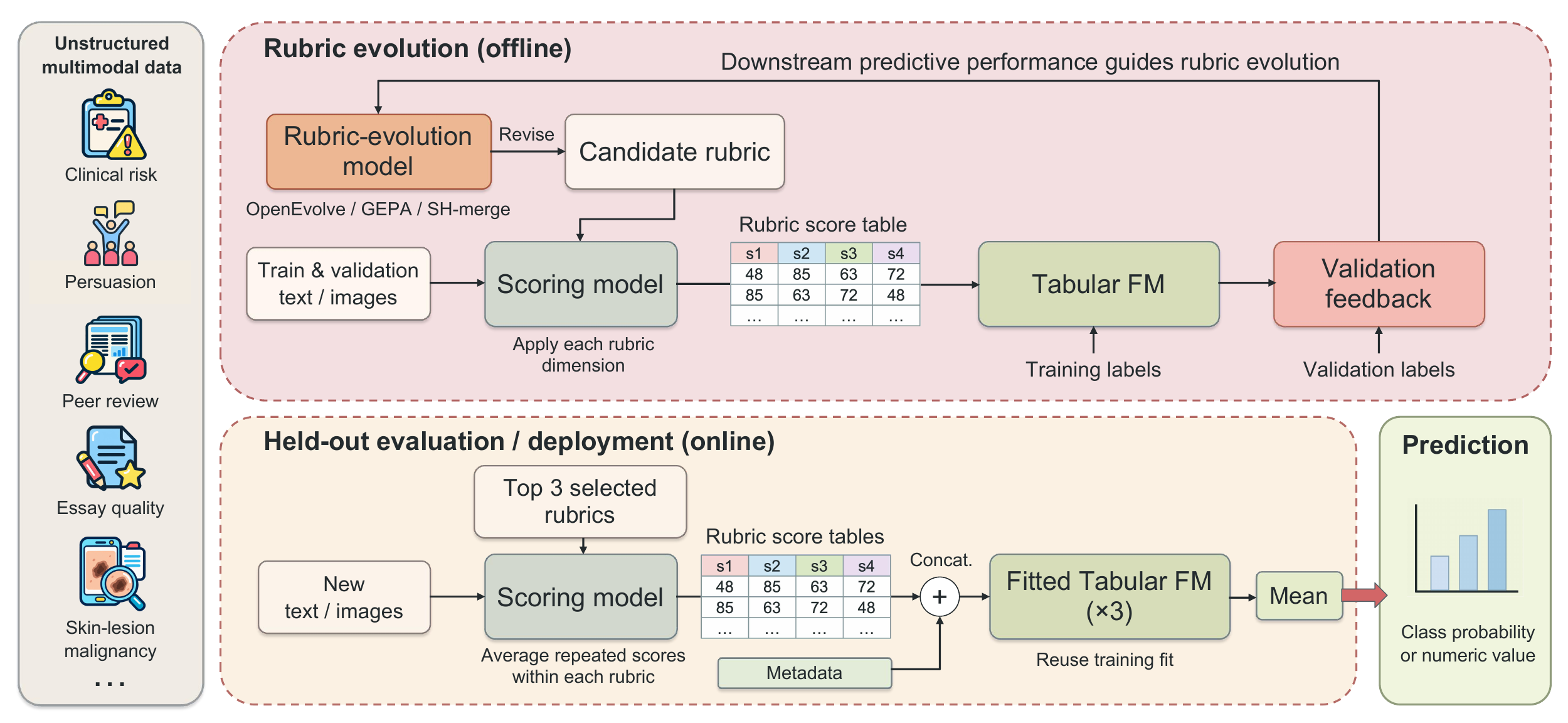}
\caption{\textbf{Overview of TableLoop.} \textbf{Top:} A language or vision-language model applies candidate rubrics to produce numeric features. A tabular predictor's validation performance and errors guide rubric revisions, without updating model weights. \textbf{Bottom:} The three rubrics selected by rubric-only validation performance produce feature tables that are combined with structured measurements. A separate tabular predictor is fitted for each rubric, and their predictions are averaged.}
\label{fig:tableloop}
\end{figure*}

\subsection{Framework Overview}

TableLoop combines unstructured inputs, such as text or images, with structured measurements by learning task-specific scoring rubrics (Figure \ref{fig:tableloop}). Each rubric defines a set of named properties and instructions for assigning numeric scores to them. A \emph{scoring model}, implemented by a language or vision language model, applies the rubric to each unstructured input, producing one feature per rubric dimension. To reduce variation between responses, we score each input $M$ times and average the scores separately for each dimension. The scoring model receives neither the instance's structured measurements nor its prediction target.

We divide the data into training, validation, and held-out test sets. Starting from an initial rubric, TableLoop alternates between evaluating candidate rubrics and proposing revisions. For each candidate, we apply a tabular foundation model to its rubric scores and training targets, then evaluate validation performance using the task-specific metric: ROC-AUC or PR-AUC for classification, and MAE for regression. The \emph{rubric-evolution model} uses validation performance and high-error examples to propose changes to the rubric's criteria and scoring instructions. The search method determines which candidates and feedback are used to generate revisions, as described below. Structured measurements are excluded throughout rubric evaluation and selection, and the test set is not used during search. Language or vision-language model weights remain fixed; optimization occurs through rubric revisions and fitting the tabular predictors.

After the search budget is exhausted, we select the three rubrics with the best rubric-only validation performance. For each selected rubric, we combine its scores with the structured measurements and apply the tabular foundation model to the test set (with only the training set as context). At test time, we average the three predictors' probabilities for classification or scalar predictions for regression:
\begin{equation}
\widehat y_i
= \frac{1}{K}\sum_{k=1}^{K}
f_k\left([s_i^{R_k};m_i]\right),
\qquad K=3,
\end{equation}
where $s_i^{R_k}$ contains the scores assigned to instance $i$ under rubric $R_k$, $m_i$ contains its structured measurements, and $f_k$ is the corresponding final predictor. The brackets denote feature concatenation. TableLoop supports interchangeable scoring models, tabular predictors, and rubric-evolution methods.

\subsection{Rubric Evolution Methods}

We implement TableLoop with OpenEvolve, GEPA, and our successive halving with complementary merge method (SH-merge). OpenEvolve maintains a population of candidate rubrics across multiple groups, or islands, together with an archive of strong candidates \citep{sharma2025openevolve}. At each iteration, the rubric-evolution model rewrites a selected parent using its validation performance, high-error examples, and search history. The revised rubric is evaluated and added to the population. Our GEPA implementation instead emphasizes detailed feedback on a small validation minibatch \citep{agrawal2025gepa}. The rubric-evolution model receives the selected rubric together with each example's input, target, rubric scores, and tabular prediction. A proposed revision is first evaluated on the same minibatch. Promising revisions receive full-validation evaluation and enter the candidate pool. Both methods run to fixed search budgets, with implementation settings given in the experimental protocol.

\textbf{SH-merge} combines parallel exploration with repeated rubric merging. We first evaluate the initial rubric, then ask the rubric-evolution model to generate $C$ variants using sampled high-error validation examples. Within each round, all candidates are evaluated on the same sampled training and validation instances and ranked by validation performance. We pair the highest-ranked candidate with the lowest-ranked, the second-highest with the second-lowest, and so on. For each pair, the rubric-evolution model receives both rubrics, their performance and validation errors, and SHAP feature importance \citep{lundberg2017shap}, then produces one child rubric by combining or revising their criteria. Thus, every parent can contribute to the next round, while the population halves. As the population shrinks, the number of instances used to evaluate each candidate doubles:
\begin{equation}
B_r=\min\!\left(2^{r-1}B,\,
N_{\mathrm{train}}+N_{\mathrm{val}}\right),
\end{equation}
where $B$ is the initial per-candidate sample budget, $r$ indexes evaluation rounds, and $N_{\mathrm{train}}$ and $N_{\mathrm{val}}$ are the split sizes. Training and validation instances retain their separate roles throughout. This schedule keeps the scorer-instance budget approximately constant per round while evaluating later candidates on more data. Evaluations can run in parallel within each round, and the search ends after a fixed number of rounds. The final round uses the full training and validation sets, and its three best rubrics form the final ensemble.

\section{Experimental Setup}

\paragraph{Datasets.}
We evaluate nine prediction tasks drawn from six datasets. Four tasks predict ICLR paper acceptance within separate research areas, using paper text and structured paper attributes. The training set uses ICLR 2025 papers, while validation and testing use ICLR 2026 papers. The remaining tasks predict essay proficiency from ELLIPSE corpus essays and writing-related measurements \citep{crossley2023ellipse}, donation outcomes from Persuasion For Good dialogues and participant personal/psychological attributes \citep{wang2019persuasion}, invasive ventilation within 24 hours of a radiology report from MIMIC-IV clinical text and measurements \citep{johnson2023mimiciv,johnson2023mimicived,johnson2023mimicivnote}, skin cancer from PAD-UFES-20 lesion images and patient and lesion attributes \citep{pacheco2020pad}, and fungal species identity from a set of five visually similar Russula species based on photographs and observation records in the Danish Fungi dataset \citep{picek2022danish}. We refer to the MIMIC-IV ventilation task as MIMIC-MV (Mechanical Ventilation) and Persuasion For Good as Persuasion below. We evaluate MAE for ELLIPSE, PR-AUC for MIMIC-MV, and ROC-AUC for the remaining tasks. The appendix describes the inputs, splits, and evaluation procedures.

\paragraph{TableLoop.}
We use Claude Sonnet 4.5 for rubric evolution, except on the MIMIC-MV task, where we use locally hosted Qwen3-235B (the data use agreement requires running only local language models). Inputs are scored with Qwen3-32B for text and Qwen3-VL-32B-Instruct for images \citep{yang2025qwen3, bai2025qwen3}, averaging three responses per instance. TabPFN is the predictor in all main experiments. All searches begin with a single-dimension rubric. OpenEvolve runs for 50 iterations, GEPA for 30 full-evaluation iterations, and SH-merge for five evaluation rounds starting from 64 candidates. Rubric evolution and selection exclude structured measurements; these are added when fitting the final three-predictor ensemble. No language or vision-language model weights are updated.

\paragraph{Baselines.}
We compare TableLoop with metadata-only TabPFN and direct language or vision-language model prediction on every task, using the same data splits. Metadata-only TabPFN predicts from the structured measurements alone. Direct prediction asks Claude Sonnet 4.5 to predict the target from the task description together with the structured and unstructured inputs, and uses locally hosted Qwen3-235B on the MIMIC-MV task. Its zero-shot predictions are then calibrated on the validation set.

\paragraph{Related feature engineering method.}
FEST (Feature Engineering with Self-Evolving Trees) \citep{khurana2026fest}, designed for binary classification, provides a closely related comparison on our seven binary tasks: like TableLoop, it converts unstructured inputs into interpretable features for downstream prediction. Specifically, FEST discovers semantic predicates and executable features from contrasting positive and negative examples, consolidates them into a shared feature bank, and iteratively prunes the bank using decision-tree importance. Because the authors do not release code, we reimplement FEST as closely as possible using the procedure, prompts, and hyperparameters specified in the paper. For a matched evaluation, we combine its selected features with the same structured measurements and use the same TabPFN predictor and data splits as TableLoop.

\section{Results}

\begin{figure*}[t]
    \centering
    \includegraphics[width=\textwidth]{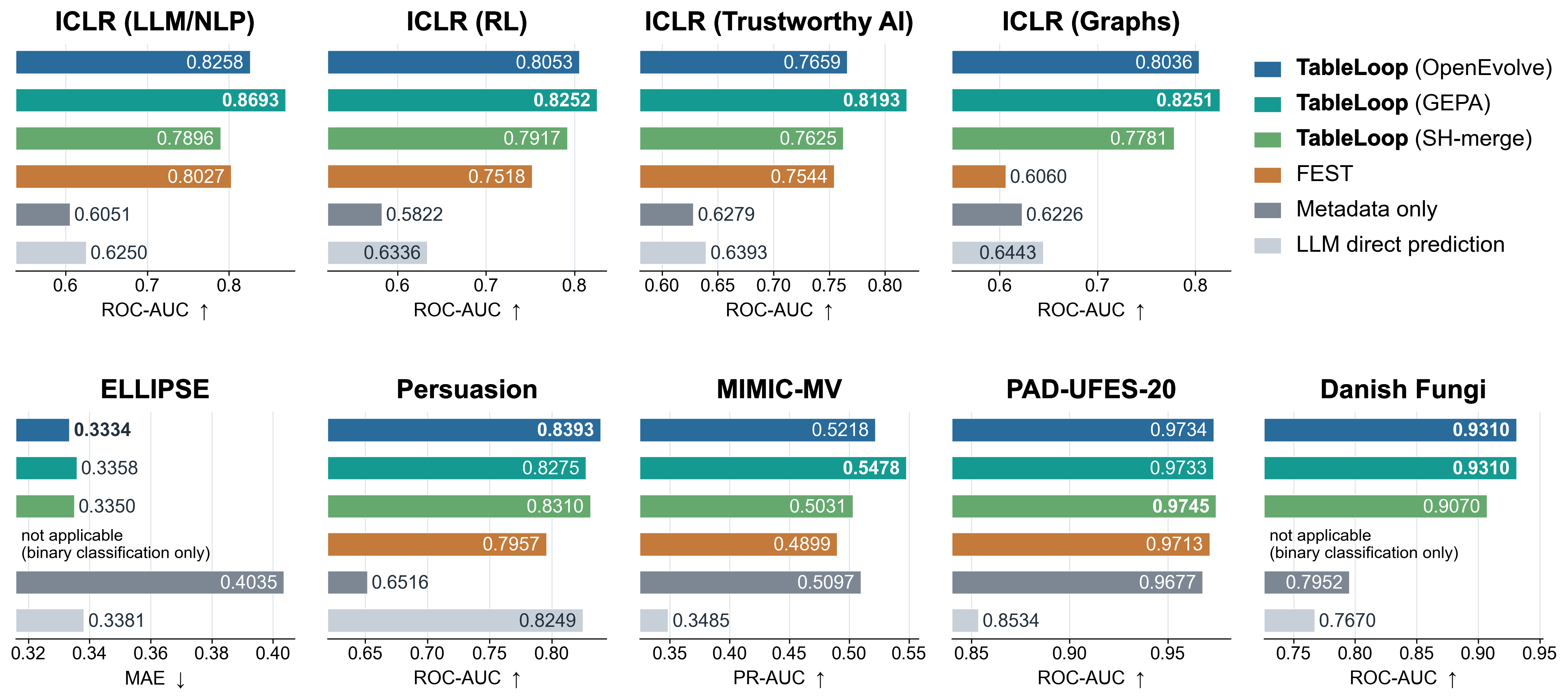}
    \caption{\textbf{TableLoop gains persist across evolution methods.}
    Results are shown for OpenEvolve, GEPA, and SH-merge using the \topthree{} selected by rubric-only validation performance. Metadata is added after selection to fit the final TabPFN predictors.
The baselines are metadata-only TabPFN and metadata-aware zero-shot direct prediction with validation-fitted affine calibration. For the seven binary classification tasks, it also reports results from FEST \citep{khurana2026fest}.
    Each panel reports the native held-out test metric: MAE for ELLIPSE, PR-AUC for MIMIC-MV, and ROC-AUC otherwise.
  }
    \label{fig:main-results}
\end{figure*}

TableLoop outperforms both structured-measurement-only TabPFN and direct prediction by a larger language or vision language model in 26 of 27 configurations across nine tasks and three evolution methods (Figure \ref{fig:main-results}). All three evolution methods exceed both baselines on eight tasks. On the MIMIC-MV prediction task, OpenEvolve and GEPA exceed both baselines, while SH-merge falls below metadata-only TabPFN. The best-performing TableLoop variant exceeds both baselines on every task.

The largest gains occur in ICLR acceptance prediction, where the best TableLoop variant improves ROC-AUC by 28.1--39.1\% relative to the stronger of the two baselines. Across the remaining tasks, the corresponding improvements are 7.5\% in MIMIC-MV PR-AUC, 1.7\% in Persuasion ROC-AUC, 0.7\% in PAD-UFES-20 lesion-level ROC-AUC, and 17.1\% in Danish Fungi observation-level macro ROC-AUC, with a 1.4\% reduction in ELLIPSE MAE. These gains span text and image inputs, tasks ranging from binary and multiclass classification to regression, and the temporal shift from ICLR 2025 training papers to ICLR 2026 evaluation papers.

TableLoop also consistently outperforms FEST across the seven binary classification tasks. OpenEvolve and GEPA outperform FEST on all seven tasks, while SH-merge does so on six, falling short only on ICLR LLM/NLP. The advantage is particularly pronounced on ICLR Graphs, where the relative improvements are $28.4$--$36.2\%$. These results support jointly evolving a complete rubric rather than constructing and pruning a bank of individually proposed features.

No single evolution method performs best on every task. GEPA leads on MIMIC-MV and all four ICLR categories, OpenEvolve on Persuasion and ELLIPSE, and SH-merge on PAD-UFES-20. The consistent gains across these methods support TableLoop as a flexible approach to combining unstructured information with structured measurements. They also show that a smaller model used to extract rubric features can support better predictions than direct use of a larger model.

Beyond the choice of evolution method, we test whether TableLoop is robust to the downstream tabular predictor by replacing TabPFN with TabFM \citep{tabfm2026} and \mbox{XGBoost~\citep{chen2016xgboost}}. TabFM performs similarly to TabPFN, and both generally outperform XGBoost (Appendix Table~\ref{tab:full-results}), suggesting that the benefit of rubric-derived representations is not specific to TabPFN and that tabular foundation models are well suited as downstream predictors in TableLoop. 

\subsection{Evolution Improves Test Performance}

To determine whether held-out performance improves as rubric evolution progresses, we conduct a retrospective truncation analysis of OpenEvolve. In the main experiments, the test set is evaluated only once after evolution is complete. For this ablation, we instead truncate the run after each iteration $k$, select the current \topthree{} using rubric-only validation performance, add metadata when applying the corresponding TabPFN predictors, and evaluate that ensemble on the test set. For ICLR, the results are averaged across three folds. While the main experiments do not use early stopping, with each OpenEvolve run using a fixed 50-iteration budget, we also retrospectively consider validation-based early stopping, where patience $p$ means that evolution stops after $p$ consecutive iterations without an improvement in the best-so-far validation score.

\begin{figure*}[t]
\centering
\includegraphics[width=\textwidth]{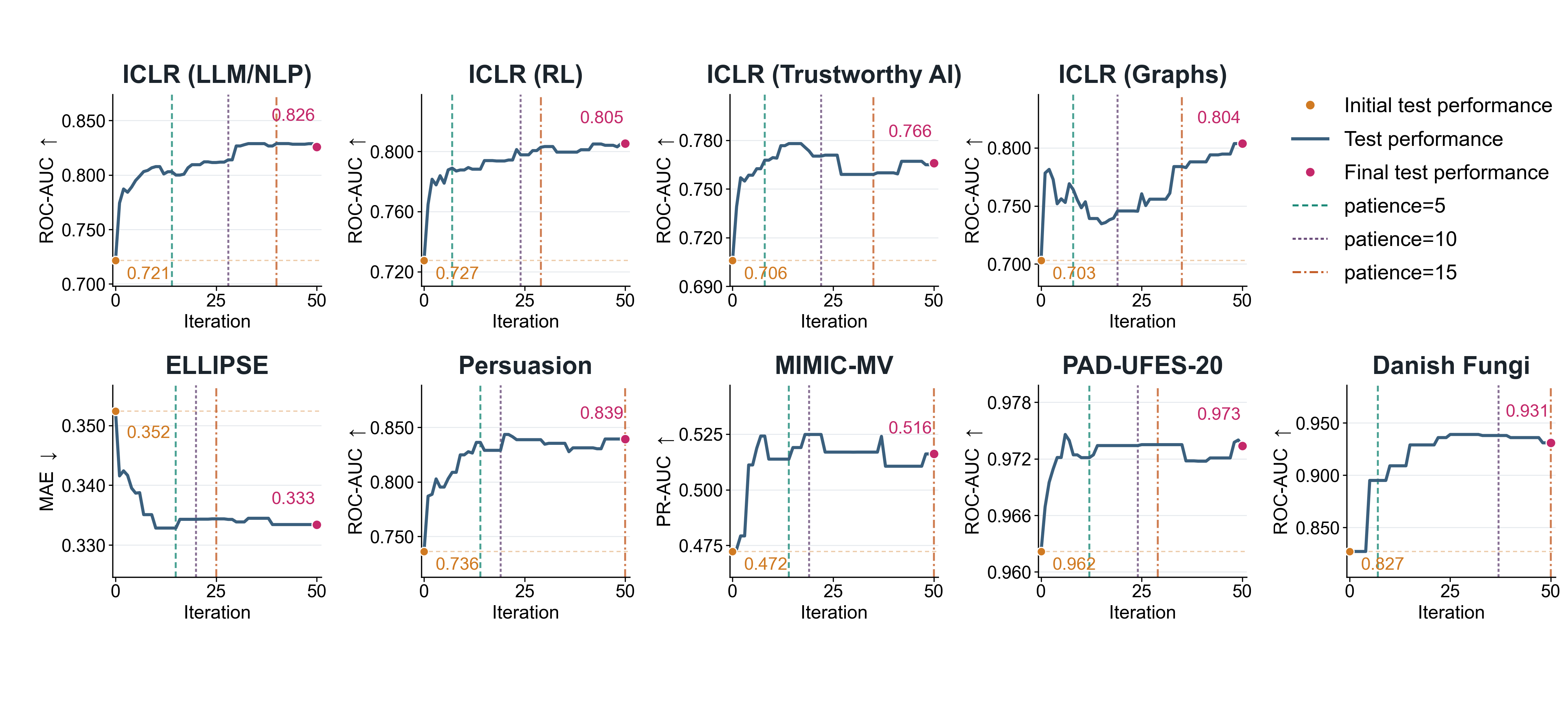}
\caption{\textbf{Evolution improves held-out test performance.}
At each iteration $k$ of a 50-iteration OpenEvolve run, the current validation-selected \topthree{} is evaluated on the test set.
Iteration 0 is the naive initial rubric, and the dotted horizontal line marks its test performance.
Vertical lines mark hypothetical early-stopping points under validation patience $5$, $10$, and $15$.}
\label{fig:oe-iteration}
\end{figure*}

Figure \ref{fig:oe-iteration} shows that held-out performance generally improves throughout evolution across all nine tasks.
The gains accumulate over multiple iterations as stronger rubrics enter the validation-selected \topthree{}, although individual trajectories include plateaus and occasional reversals.
The final ensembles achieve relative improvements of up to 14.5$\%$, showing that improvements selected on the validation set generally carry over to held-out test data.
As a retrospective analysis, the vertical lines in Figure \ref{fig:oe-iteration} mark where the validation-based early-stopping rule would have halted.
If we introduced this form of early stopping, it would often reach a substantial fraction of the test gain at a smaller evaluation cost, but it can also halt too soon and miss later improvements.

The main experiments initialize evolution from a deliberately minimal, single-dimension rubric. To examine whether a more informative starting point improves the search, we additionally compare this naive initialization with expert-designed, task-specific rubrics for ICLR and ELLIPSE datasets. The naive initialization performs better across all five evaluated tasks, suggesting that starting from a broad rubric leaves greater flexibility for evolution to discover useful task-specific dimensions. Full details and results are provided in Appendix~\ref{sec:init-abla}.

\subsection{Instance-Level Validation Feedback}

TableLoop uses two forms of validation feedback during evolution: an aggregate validation metric and examples on which the current predictor makes large errors. To isolate the contribution of the latter, we rerun OpenEvolve while withholding the validation examples, labels, rubric scores, and predictions associated with these errors. The evolution model still receives the aggregate validation metric; all other settings remain unchanged.

\begin{table}[H]
\centering
\small
\caption{\textbf{Instance-level validation feedback improves rubric evolution.}
OpenEvolve results with the standard feedback and with only the aggregate validation metric. Results use the validation-selected \topthree{} aggregation. ICLR results are averaged across three folds.}
\label{tab:validation-feedback}
\begin{tabular}{llcc}
\toprule
Task & Metric & Instance feedback & Without Instance feedback \\
\midrule
ICLR (LLM/NLP)       & ROC-AUC $\uparrow$       & 0.8258 & 0.8055 \\
ICLR (RL)            & ROC-AUC $\uparrow$       & 0.8053 & 0.8044 \\
ICLR (Trustworthy AI)& ROC-AUC $\uparrow$       & 0.7659 & 0.7318 \\
ICLR (Graphs)        & ROC-AUC $\uparrow$       & 0.8036 & 0.7653 \\
ELLIPSE               & MAE $\downarrow$         & 0.3334 & 0.3347 \\
Persuasion            & ROC-AUC $\uparrow$       & 0.8393 & 0.7747 \\
MIMIC-MV   & PR-AUC $\uparrow$        & 0.5218 & 0.2589 \\
PAD-UFES-20           & ROC-AUC $\uparrow$       & 0.9734 & 0.9686 \\
Danish Fungi          & ROC-AUC $\uparrow$ & 0.9310 & 0.9179 \\
\bottomrule
\end{tabular}
\end{table}

Withholding instance-level error feedback generally degrades held-out performance, even though the evolution model still observes the aggregate validation metric. The metric indicates whether a rubric performs well overall, but does not identify which distinctions the rubric fails to capture or score correctly. These results show that instance-level validation feedback is important for directing useful rubric revisions.

\subsection{Does Evolution Discover Complementary Rubrics?}

We examine whether rubric evolution expands the set of examples that can be predicted correctly and discovers rubrics with complementary strengths. To distinguish this from simply improving predictions on examples already handled well, we measure \emph{coverage}: the fraction of validation instances for which at least one rubric-specific predictor produces a correct prediction. For classification, correctness uses a probability threshold of 0.5; for ELLIPSE, it requires an absolute error of at most $\epsilon$.

Using rubric-only validation predictions from 50 OpenEvolve iterations, we compare four selections: the initial rubric (Ini.), the evolved rubric with the highest individual coverage (Cov-1), the three rubrics with the highest combined coverage (Cov-3), and the standard three rubrics selected by validation task performance (Top-3). Ini.\ versus Cov-1 tests whether evolution finds individual rubrics that correctly predict more examples, while comparing Cov-1 with the three-rubric selections tests whether different rubrics succeed on different examples.

Coverage increases from Ini.\ to Cov-1 on every task (Table \ref{tab:coverage}), showing that evolution discovers individual rubrics that correctly predict a larger fraction of validation examples. Top-3 further exceeds Cov-1 on every task, indicating that the selected rubrics have complementary successes rather than predicting the same examples correctly. For example, on ICLR LLM/NLP, coverage rises from 58.7$\%$ for Ini.\ to 68.4$\%$ for Cov-1 and 84.0$\%$ for Top-3 before adding metadata.

We also test whether explicitly selecting for coverage improves the final ensemble. Cov-3 achieves the highest validation coverage by construction, but its averaged predictions yield worse held-out performance than Top-3 on six of nine tasks. Having at least one correct predictor for an example does not guarantee that averaging the predictors will improve the final prediction. These results support using coverage to assess complementary strengths while retaining validation task performance as the default rubric-selection criterion.
\begin{table}[H]
\centering
\scriptsize
\caption{\textbf{Evolution discovers rubrics with complementary predictive strengths.} Validation coverage is the fraction of examples predicted correctly by at least one selected rubric-specific predictor, using rubric scores alone. Ini.\ uses the initial rubric; Cov-1 selects the individual rubric with highest coverage; Cov-3 selects three rubrics by their combined coverage; and Top-3 selects three by validation task performance. The final two columns report held-out performance after adding structured measurements and averaging predictions. ICLR results are averaged across three folds.}
\label{tab:coverage}
\resizebox{\columnwidth}{!}{%
\begin{tabular}{llcccccc}
\toprule
& & \multicolumn{4}{c}{Val.\ coverage} & \multicolumn{2}{c}{Test metric} \\
\cmidrule(lr){3-6} \cmidrule(lr){7-8}
Task & Metric & Ini. & Cov-1 & Cov-3 & Top-3 & Cov-3 & Top-3 \\
\midrule
ICLR (LLM/NLP) & ROC-AUC \metricup & 0.5865 & 0.6837 & 0.8646 & 0.8404 & 0.8187 & 0.8258 \\
ICLR (RL) & ROC-AUC \metricup & 0.6189 & 0.6831 & 0.8811 & 0.8207 & 0.7907 & 0.8053 \\
ICLR (Trustworthy AI) & ROC-AUC \metricup & 0.5597 & 0.6511 & 0.8573 & 0.8083 & 0.7789 & 0.7659 \\
ICLR (Graphs) & ROC-AUC \metricup & 0.6903 & 0.7308 & 0.9365 & 0.8693 & 0.7960 & 0.8036 \\
ELLIPSE ($\epsilon=.5$) & MAE \metricdown & 0.7045 & 0.7521 & 0.8473 & 0.8304 & 0.3352 & 0.3334 \\
Persuasion & ROC-AUC \metricup & 0.6483 & 0.7517 & 0.8690 & 0.8138 & 0.8208 & 0.8393 \\
MIMIC-MV & PR-AUC \metricup & 0.8527 & 0.8549 & 0.8637 & 0.8615 & 0.5252 & 0.5218 \\
PAD-UFES-20 & ROC-AUC \metricup & 0.6736 & 0.8012 & 0.9258 & 0.8754 & 0.9740 & 0.9734 \\
Danish Fungi & ROC-AUC \metricup & 0.3980 & 0.7245 & 0.8980 & 0.8061 & 0.9208 & 0.9310 \\

\bottomrule
\end{tabular}}
\end{table}

\subsection{Complementarity with Metadata}

For some tasks, the available metadata is comprehensive: it includes attributes closely related to the unstructured multimodal data and achieves strong predictive performance on its own.
To examine the contribution of TableLoop when this overlap is reduced, we conduct a controlled ablation on PAD-UFES-20.
We remove nine potentially overlapping lesion-specific measurements and symptoms: region, two diameter measurements, elevation, itching, growth, pain, change, and bleeding, while retaining 12 patient-level demographic, background, and risk-history fields.
The unstructured input, selected rubrics, and TabPFN configuration are held fixed; only the metadata set supplied to TableLoop and its matched metadata-only baseline changes.

\begin{table}[H]
\centering
\small
\caption{Controlled metadata ablation on PAD-UFES-20, measured by test lesion-level ROC-AUC. The reduced setting removes nine lesion-specific fields while retaining 12 patient-level demographic, background, and risk-history fields.}
\label{tab:metadata-ablation}
\begin{tabular}{lcccc}
\toprule
Metadata set & Fields & Metadata only & TableLoop & TableLoop gain \\
\midrule
Full & 21 & 0.9677 & 0.9734 & $+0.0057$ \\
Patient-level only & 12 & 0.8709 & 0.9345 & $+0.0636$ \\
\bottomrule
\end{tabular}
\end{table}

The gain from TableLoop over metadata-only prediction increases from $0.0057$ with full metadata to $0.0636$ after the potentially overlapping fields are removed.
This larger gap shows that TableLoop extracts useful predictive information from the unstructured multimodal data and that its rubric features complement the remaining structured metadata.

\begingroup
\color{revisionblue}
\subsection{Comparison with Frozen Embeddings}
\label{sec:embedding-comparison}

We compare our TableLoop method with embedding baselines on seven tasks. Rather than generating features from a rubric using a scoring model, our embedding baselines pass to TabPFN the text embeddings along with the relevant metadata. We  use Qwen3-Embedding-8B \citep{zhang2025qwen3} for text and Qwen3-VL-Embedding-8B \citep{li2026qwen3} for images. We evaluate predictions using the original 4,096-dimensional embeddings and a reduced version obtained using principal component analysis (PCA). PCA is fit only on the training embeddings, retaining at least 95\% of their variance. For each representation, TabPFN takes either the embeddings alone or the embeddings concatenated with the same metadata used by TableLoop. All methods use identical data splits and evaluation metrics. For both ICLR tasks, all methods use the same anonymized paper bodies from PaperScorer-Text \citep{konan2026paperscorer} (see Appendix~\ref{app:paperscorer}).

Table~\ref{tab:embedding-comparison} shows that TableLoop outperforms all four embedding variants on six of the seven tasks, including all five text tasks. Its advantage is most pronounced on ICLR (RL) and Persuasion, where it improves ROC-AUC by 0.1094 and 0.0690, respectively, over the strongest embedding baseline for each task. We hypothesize that the gains over pure embedding methods are larger on tasks like paper reviewing because they are inherently complex tasks that deal with larger bodies of text and more complex reasoning. In contrast, our image classification tasks are simpler and so there is potentially less to be gained from creating a thoughtful rubric and spending inference compute reasoning about what scores should be given. Our results suggest that task-adapted rubric scores can be more useful to tabular foundation models for predicting the target than general-purpose frozen embeddings.

\begin{table}[H]
\centering
\small
\color{revisionblue}
\caption{\textbf{TableLoop outperforms frozen embedding baselines on 6 of 7 tasks.}
All methods use TabPFN. TableLoop results are for the OpenEvolve variant.
ICLR results are averaged across three folds.}
\label{tab:embedding-comparison}
\begin{tabular}{llccccc}
\toprule
& & \multicolumn{2}{c}{Embedding} & \multicolumn{2}{c}{Embedding + metadata} & \\
\cmidrule(lr){3-4} \cmidrule(lr){5-6}
Task & Metric & 4096 & PCA 95\% & 4096 & PCA 95\% & TableLoop \\
\midrule
Persuasion    & ROC-AUC $\uparrow$ & 0.7446 & 0.6946 & 0.7703 & 0.7093 & \textbf{0.8393} \\
ELLIPSE       & MAE $\downarrow$   & 0.3927 & 0.3925 & 0.3859 & 0.3743 & \textbf{0.3334} \\
MIMIC-MV      & PR-AUC $\uparrow$  & 0.4281 & 0.4126 & 0.4205 & 0.4791 & \textbf{0.5218} \\
ICLR (RL)     & ROC-AUC $\uparrow$ & 0.5891 & 0.5651 & 0.5888 & 0.5920 & \textbf{0.7014} \\
ICLR (Graphs) & ROC-AUC $\uparrow$ & 0.6371 & 0.6084 & 0.6546 & 0.6238 & \textbf{0.6946} \\
PAD-UFES-20   & ROC-AUC $\uparrow$ & 0.9252 & 0.9185 & 0.9655 & \textbf{0.9743} & 0.9734 \\
Danish Fungi  & ROC-AUC $\uparrow$ & 0.8885 & 0.9052 & 0.8940 & 0.9198 & \textbf{0.9310} \\
\bottomrule
\end{tabular}
\end{table}

\endgroup

\section{Discussion}

TableLoop provides a flexible framework for combining structured measurements with information from text and images. By learning what a language or vision-language model should measure, it turns unstructured inputs into explicit features that tabular predictors can use alongside existing data. Predictive feedback guides revisions to the rubric's criteria and scoring instructions, allowing the representation to adapt to the task without finetuning the language model.

Across nine tasks, TableLoop outperforms both structured-data-only learning and calibrated direct prediction by a larger language or vision-language model, showing that a smaller model used to extract task-specific features can support stronger predictions than direct use of a larger model. The ablations further show that evolution can improve held-out performance and discover rubrics with complementary predictive strengths, while the resulting features remain useful across different tabular predictors. Although the learned rubrics are inspectable, their alignment with the intended concepts warrants further human evaluation. More broadly, as with other validation-guided optimization methods, studies across additional datasets, data splits, and scorer LLMs would help assess robustness.

{\color{revisionblue}The comparison with frozen embeddings further shows that explicitly
task-adapted rubric features can be more predictive than general-purpose
representations. A next step could be to explore the joint use of frozen embeddings and rubric
scores with structured measurements, and to assess whether the two
representations capture complementary predictive information.}

Efficiency is an important consideration because rubric evolution requires repeatedly scoring examples and evaluating candidate representations, with costs depending on the dataset size, input length, and search budget. This optimization takes place offline: once the rubrics are selected, predictions on new inputs require only the fixed scoring procedure and downstream tabular predictors. {\color{revisionblue} Several strategies could reduce this offline cost. Early stopping
would reduce the number of candidates evaluated, while training-set subsampling and fewer scoring repetitions would reduce the number of calls per candidate. In our ELLIPSE experiments, subsampling the training set while retaining the full validation set reduced scoring calls per iteration while maintaining comparable predictive performance (Appendix~\ref{sec:train-subsampling}). For the calls that remain, parallel execution can shorten wall-clock time during both search and prediction. The scorer itself offers another opportunity to reduce costs. On the same rubrics, JEV System~1 (9B) achieves a roughly $5\times$ speedup over the similarly sized Qwen3-8B, measured in GPU time per example (Table~\ref{tab:gpu-seconds}), highlighting the potential of System~1
models for efficient rubric scoring. Future work could explore further speedups by scoring multiple rubric
dimensions in parallel in a single forward pass, and assess JEV's
impact on TableLoop's end-to-end runtime and predictive performance.}



Beyond its final predictions, TableLoop produces reusable feature tables grounded in real observations. A learned rubric can be applied to additional inputs, and its scores can be stored, inspected, and used with other predictors. This creates opportunities to convert labeled text and image collections from different domains into tabular datasets, potentially expanding the data available for training and adapting tabular foundation models. Although this application remains to be evaluated, it highlights the broader utility of the framework: learning explicit, task-relevant representations that make unstructured information accessible to tabular learning.

\subsection*{Acknowledgments}
This work was supported in part by the National Science Foundation under grants CCF-2550179, CCF-2403075, and CCF-2212426; by the Office of Naval Research under grant N000142412289; by an Amazon Research Award, and a gift by Coefficient Giving.

\subsection*{AI use statement}

We used generative AI tools to edit the manuscript for readability and to aid retrieval and discovery of related work. The language models used as scorers and evolution engines are part of the proposed method and experimental pipeline, not authoring assistants. We have reviewed all AI-assisted work: authors checked edited text against the paper's methods and results, and verified suggested citations against the original sources. We take responsibility for the final content of this work, including text, claims or artifacts produced with the aid of generative AI.

\subsection*{Ethics statement}

We access MIMIC-IV, MIMIC-IV-ED, and MIMIC-IV-Note under their data-use requirements. These deidentified clinical records remain in our local computing environment, and all language-model processing of them uses locally hosted models; we do not send them to external model providers or redistribute them. The other datasets contain potentially sensitive human-generated text, demographic attributes, or clinical images. We use only the released data for the stated research purposes, and make no attempt to identify participants.
\subsection*{Reproducibility statement}

The main text describes the TableLoop pipeline, rubric evolution and selection, baselines, and evaluation metrics. Appendix \ref{app:experimental-setup} provides additional details on dataset construction and splits, input features, models, hyperparameters, search budgets, missing-value handling, and baseline configurations. The appendix also reports extended baseline and ablation results. MIMIC data must be obtained independently through its authorized access process; the other datasets can be obtained from their original sources.

\bibliography{iclr2027_conference}
\bibliographystyle{iclr2027_conference}
\clearpage
\appendix
\section{Downstream Predictor Results}

Although the main experiments use TabPFN, TableLoop is not tied to a specific downstream tabular predictor. We compare TabPFN with TabFM and XGBoost while holding the selected \topthree{} rubrics and input features fixed for each evolution method. Table~\ref{tab:full-results} reports the results across OpenEvolve, GEPA, and SH-merge.
\begin{table}[H]
\centering
\scriptsize
\setlength{\tabcolsep}{2.8pt}
\caption{\textbf{Complete evolution-method-by-backbone results.}
Each dataset uses its native metric.
OpenEvolve uses a 50-iteration budget; GEPA uses 30 full-evaluation iterations.
``Direct prediction'' denotes the metadata-aware zero-shot direct prediction score; ``Metadata-only'' uses only metadata; ``TableLoop'' denotes the default method with its validation-selected \topthree{} and metadata.}
\label{tab:full-results}
\resizebox{\textwidth}{!}{%
\begin{tabular}{ll|ccc|ccc|ccc}
\toprule
Dataset & Setting &
\multicolumn{3}{c|}{OpenEvolve} &
\multicolumn{3}{c|}{GEPA} &
\multicolumn{3}{c}{SH-merge} \\
\cmidrule(lr){3-5}\cmidrule(lr){6-8}\cmidrule(lr){9-11}
& & PFN & TabFM & XGB & PFN & TabFM & XGB & PFN & TabFM & XGB \\
\midrule

\multirow{3}{*}{\shortstack[l]{ICLR (LLM/NLP)\\ROC-AUC \metricup}}
& Direct prediction & \multicolumn{9}{c}{0.6250} \\
& Metadata-only & 0.6051 & 0.6150 & 0.5476 & 0.6051 & 0.6150 & 0.5476 & 0.6051 & 0.6150 & 0.5476 \\
& TableLoop & 0.8258 & 0.8301 & 0.7980 & 0.8693 & \better\textbf{0.8721} & 0.8493 & 0.7896 & 0.7914 & 0.7697 \\
\midrule

\multirow{3}{*}{\shortstack[l]{ICLR (RL)\\ROC-AUC \metricup}}
& Direct prediction & \multicolumn{9}{c}{0.6336} \\
& Metadata-only & 0.5822 & 0.5787 & 0.5510 & 0.5822 & 0.5787 & 0.5510 & 0.5822 & 0.5787 & 0.5510 \\
& TableLoop & 0.8053 & 0.8059 & 0.7804 & 0.8252 & \better\textbf{0.8257} & 0.7925 & 0.7917 & 0.7919 & 0.7818 \\
\midrule

\multirow{3}{*}{\shortstack[l]{ICLR (Trustworthy AI)\\ROC-AUC \metricup}}
& Direct prediction & \multicolumn{9}{c}{0.6393} \\
& Metadata-only & 0.6279 & 0.6327 & 0.5512 & 0.6279 & 0.6327 & 0.5512 & 0.6279 & 0.6327 & 0.5512 \\
& TableLoop & 0.7659 & 0.7625 & 0.7394 & 0.8193 & \better\textbf{0.8196} & 0.8042 & 0.7625 & 0.7586 & 0.7420 \\
\midrule

\multirow{3}{*}{\shortstack[l]{ICLR (Graphs)\\ROC-AUC \metricup}}
& Direct prediction & \multicolumn{9}{c}{0.6443} \\
& Metadata-only & 0.6226 & 0.5917 & 0.5367 & 0.6226 & 0.5917 & 0.5367 & 0.6226 & 0.5917 & 0.5367 \\
& TableLoop & 0.8036 & 0.7936 & 0.8049 & \better\textbf{0.8251} & 0.8238 & 0.8039 & 0.7781 & 0.7681 & 0.7636 \\
\midrule

\multirow{3}{*}{\shortstack[l]{ELLIPSE\\MAE \metricdown}}
& Direct prediction & \multicolumn{9}{c}{0.3381} \\
& Metadata-only & 0.4035 & 0.4034 & 0.4227 & 0.4035 & 0.4034 & 0.4227 & 0.4035 & 0.4034 & 0.4227 \\
& TableLoop & 0.3334 & \better\textbf{0.3315} & 0.3427 & 0.3358 & 0.3345 & 0.3431 & 0.3350 & 0.3330 & 0.3417 \\
\midrule

\multirow{3}{*}{\shortstack[l]{Persuasion\\ROC-AUC \metricup}}
& Direct prediction  & \multicolumn{9}{c}{0.8249} \\
& Metadata-only & 0.6516 & 0.6342 & 0.5519 & 0.6516 & 0.6342 & 0.5519 & 0.6516 & 0.6342 & 0.5519 \\
& TableLoop & \better\textbf{0.8393} & 0.8301 & 0.8113 & 0.8275 & 0.8328 & 0.8088 & 0.8310 & 0.8229 & 0.8133 \\
\midrule

\multirow{3}{*}{\shortstack[l]{MIMIC-MV\\PR-AUC \metricup}}
& Direct prediction & \multicolumn{9}{c}{0.3485} \\
& Metadata-only & 0.5097 & 0.4964 & 0.4219 & 0.5097 & 0.4964 & 0.4219 & 0.5097 & 0.4964 & 0.4219 \\
& TableLoop & 0.5218 & 0.5151 & 0.4646 & \better\textbf{0.5478} & 0.5255 & 0.4538 & 0.5031 & 0.4930 & 0.4591 \\
\midrule

\multirow{3}{*}{\shortstack[l]{PAD-UFES-20\\ROC-AUC \metricup}}
& Direct prediction & \multicolumn{9}{c}{0.8534} \\
& Metadata-only & 0.9677 & 0.9693 & 0.9577 & 0.9677 & 0.9693 & 0.9577 & 0.9677 & 0.9693 & 0.9577 \\
& TableLoop & 0.9734 & 0.9750 & 0.9706 & 0.9733 & 0.9758 & 0.9740 & 0.9745 & \better\textbf{0.9772} & 0.9737 \\
\midrule

\multirow{3}{*}{\shortstack[l]{Danish Fungi\\ ROC-AUC \metricup}}
& Direct prediction & \multicolumn{9}{c}{0.7670} \\
& Metadata-only & 0.7952 & 0.8009 & 0.8103 & 0.7952 & 0.8009 & 0.8103 & 0.7952 & 0.8009 & 0.8103 \\
& TableLoop & 0.9310 & \better\textbf{0.9373} & 0.9322 & 0.9310 & 0.9372 & 0.9327 & 0.9070 & 0.9157 & 0.9125 \\

\bottomrule
\end{tabular}}
\end{table}

\section{Direct LM/VLM Prediction Results}

We sweep shot count, affine calibration, and the presence of metadata to choose a strong direct-prediction baseline for comparison with TableLoop.
Scores are aggregated by aligning instances across three seeds, averaging the per-instance score, and computing the metric once.
An affine map is fitted on validation predictions and applied unchanged to test.
Several zero-shot runs are deterministic and therefore use a single seed.

Few-shot prompting substantially improves uncalibrated predictions on ELLIPSE, but these gains largely disappear after validation-fitted affine calibration. For example, with metadata, Claude Sonnet 4.5's raw MAE decreases from 0.6105 with zero shots to 0.4671 with ten shots, whereas calibrated zero-shot prediction achieves a lower MAE than any tested few-shot setting (0.3381 versus 0.3500--0.3528). The same pattern holds for Qwen3-32B, while Claude without metadata shows only small, inconsistent gains from demonstrations after calibration. This suggests that much of the benefit of demonstrations on ELLIPSE concerns matching the target score scale, which calibration can address directly. Across classification tasks, few-shot examples sometimes improve performance, but their effects vary by task and model. We therefore use calibrated zero-shot prediction as a consistent direct-prediction baseline, while reporting the few-shot comparisons here. These findings also motivate exploring ways to use labeled data beyond including examples in the prediction prompt. TableLoop uses predictive feedback to revise what the scoring model measures and trains a downstream predictor to combine the resulting features with structured measurements.

\begin{table}[H]
\centering
\small
\caption{ELLIPSE direct prediction scores (MAE \metricdown).}
\label{tab:direct-ellipse}
\begin{tabular}{lc cc cc}
\toprule
& & \multicolumn{2}{c}{Without metadata} & \multicolumn{2}{c}{With metadata} \\
\cmidrule(lr){3-4}\cmidrule(lr){5-6}
Model & Shot & Raw & Affine & Raw & Affine \\
\midrule
Claude Sonnet 4.5 & 0 & 0.6316 & 0.3431 & 0.6105 & 0.3381 \\
& 3 & 0.4679 & 0.3428 & 0.5017 & 0.3528 \\
& 5 & 0.4566 & 0.3413 & 0.4885 & 0.3524 \\
& 10 & 0.4648 & 0.3480 & 0.4671 & 0.3500 \\
\midrule
Qwen3-32B & 0 & 0.5983 & 0.3928 & 0.5727 & 0.3831 \\
& 3 & 0.3985 & 0.4045 & 0.4166 & 0.4090 \\
& 5 & 0.3952 & 0.4042 & 0.4130 & 0.4089 \\
& 10 & 0.3986 & 0.4053 & 0.4067 & 0.4107 \\
\bottomrule
\end{tabular}
\end{table}

\begin{table}[H]
\centering
\small
\caption{Persuasion direct prediction scores (ROC-AUC \metricup).}
\label{tab:direct-pers}
\begin{tabular}{lc cc cc}
\toprule
& & \multicolumn{2}{c}{Without metadata} & \multicolumn{2}{c}{With metadata} \\
\cmidrule(lr){3-4}\cmidrule(lr){5-6}
Model & Shot & Raw & Affine & Raw & Affine \\
\midrule
Claude Sonnet 4.5 & 0 & 0.7923 & 0.7923 & 0.8249 & 0.8249 \\
& 3 & 0.7268 & 0.7268 & 0.7692 & 0.7692 \\
& 5 & 0.8129 & 0.8129 & 0.8019 & 0.8018 \\
& 10 & 0.7912 & 0.7908 & 0.8033 & 0.8033 \\
\midrule
Qwen3-32B & 0 & 0.7610 & 0.7610 & 0.7667 & 0.7667 \\
& 3 & 0.7807 & 0.7804 & 0.7572 & 0.7572 \\
& 5 & 0.7952 & 0.7952 & 0.7628 & 0.7632 \\
& 10 & 0.7652 & 0.7652 & 0.7593 & 0.7594 \\
\bottomrule
\end{tabular}
\end{table}

\begin{table}[H]
\centering
\small
\caption{MIMIC-MV direct prediction scores (PR-AUC \metricup). Affine maps are monotone here, so ranking metrics are unchanged except for occasional clipping or ties.}
\label{tab:direct-medical}
\begin{tabular}{lc cc cc}
\toprule
& & \multicolumn{2}{c}{Without metadata} & \multicolumn{2}{c}{With metadata} \\
\cmidrule(lr){3-4}\cmidrule(lr){5-6}
Model & Shot & Raw & Affine & Raw & Affine \\
\midrule
Qwen3-235B & 0 & 0.2721 & 0.2721 & 0.3485 & 0.3485 \\
& 3 & 0.2908 & 0.2908 & 0.3307 & 0.3307 \\
& 5 & 0.3110 & 0.3110 & 0.3400 & 0.3399 \\
& 10 & 0.3063 & 0.3063 & 0.3315 & 0.3315 \\
\midrule
Qwen3-32B & 0 & 0.2465 & 0.2465 & 0.3001 & 0.3001 \\
& 3 & 0.2613 & 0.2613 & 0.3128 & 0.3128 \\
& 5 & 0.2809 & 0.2809 & 0.3244 & 0.3244 \\
& 10 & 0.3032 & 0.3032 & 0.3168 & 0.3168 \\
\bottomrule
\end{tabular}
\end{table}

\begin{table}[H]
\centering
\small
\caption{PAD-UFES-20 direct prediction score (lesion-level ROC-AUC \metricup). The selected Claude Sonnet 4.5 zero-shot setting includes the image and metadata.}
\label{tab:direct-pad}
\begin{tabular}{lccc}
\toprule
Model & Shot & Raw & Affine \\
\midrule
Claude Sonnet 4.5 & 0 & 0.8537 & 0.8534 \\
\bottomrule
\end{tabular}
\end{table}

\begin{table}[H]
\centering
\small
\setlength{\tabcolsep}{4pt}
\caption{ICLR direct prediction without metadata (ROC-AUC \metricup). Each cell is the mean of three fold-level metrics; within a fold, scores are averaged across seeds before the metric is computed.}
\label{tab:direct-iclr}
\begin{tabular}{llccc}
\toprule
Category & Model & Shot & Raw & Affine \\
\midrule
LLM/NLP & Claude Sonnet 4.5 & 0 & 0.6532 & 0.6532 \\
& & 3 & 0.6473 & 0.6478 \\
& & 5 & 0.6336 & 0.6335 \\
& & 10 & 0.6668 & 0.6669 \\
& Qwen3-32B & 0 & 0.6447 & 0.6447 \\
& & 3 & 0.6236 & 0.6236 \\
& & 5 & 0.6365 & 0.6365 \\
& & 10 & 0.6320 & 0.6320 \\
\midrule

RL & Claude Sonnet 4.5 & 0 & 0.6622 & 0.6622 \\
& & 3 & 0.6640 & 0.6641 \\
& & 5 & 0.6517 & 0.6514 \\
& & 10 & 0.6718 & 0.6717 \\
& Qwen3-32B & 0 & 0.6955 & 0.6955 \\
& & 3 & 0.6825 & 0.6826 \\
& & 5 & 0.6714 & 0.6709 \\
& & 10 & 0.6864 & 0.6865 \\
\midrule
Trustworthy AI & Claude Sonnet 4.5 & 0 & 0.6680 & 0.6680 \\
& & 3 & 0.6433 & 0.6442 \\
& & 5 & 0.6520 & 0.6521 \\
& & 10 & 0.6664 & 0.6661 \\
& Qwen3-32B & 0 & 0.6129 & 0.6129 \\
& & 3 & 0.5796 & 0.5795 \\
& & 5 & 0.6012 & 0.6011 \\
& & 10 & 0.5901 & 0.5901 \\
\midrule

Graphs & Claude Sonnet 4.5 & 0 & 0.6803 & 0.6803 \\
& & 3 & 0.7184 & 0.7178 \\
& & 5 & 0.6378 & 0.6383 \\
& & 10 & 0.6303 & 0.6297 \\
& Qwen3-32B & 0 & 0.5761 & 0.5761 \\
& & 3 & 0.5576 & 0.5576 \\
& & 5 & 0.6314 & 0.6314 \\
& & 10 & 0.6106 & 0.6106 \\

\bottomrule
\end{tabular}
\end{table}

\begin{table}[H]
\centering
\small
\caption{ICLR metadata-aware Claude Sonnet 4.5 zero-shot direct prediction (ROC-AUC \metricup), the setting used in the main results. Scores are means of three fold-level metrics.}
\label{tab:direct-iclr-meta}
\begin{tabular}{lcc}
\toprule
Category & Raw & Affine \\
\midrule
LLM/NLP & 0.6250 & 0.6250 \\
RL & 0.6336 & 0.6336 \\
Trustworthy AI & 0.6393 & 0.6393 \\
Graphs & 0.6443 & 0.6443 \\
\bottomrule
\end{tabular}
\end{table}

Across these grids, few-shot demonstrations do not consistently improve over shot~0, affine calibration is essential for ELLIPSE MAE while leaving ranking metrics essentially unchanged, and providing metadata typically helps.
The metadata-aware, affine-calibrated zero-shot setting is therefore the strongest overall protocol, and we use it as the direct-prediction baseline in the main experiments when comparing against TableLoop.

\subsection{Adding a Direct Prediction Score}
We finally investigate whether a direct prediction score from a frontier model can provide complementary signal and further improve TableLoop.
We consider two integration strategies: \emph{feature concatenation}, which supplies the calibrated direct score to TabPFN alongside the rubric features and metadata, and \emph{score fusion}, which fits a validation-trained ridge model over the TableLoop and direct predictions.

\begin{table}[H]
\centering
\small
\caption{Direct-prediction-score augmentation for the 50-iteration OpenEvolve \topthree{} with TabPFN and metadata. Direct scores are produced by Claude Sonnet 4.5, except on MIMIC-MV, where privacy restrictions require Qwen3-235B. ICLR results are averaged over three folds.}
\label{tab:direct-fusion}
\begin{tabular}{llccc}
\toprule
Task & Metric & TableLoop & Feature concat. & Score fusion \\
\midrule
ICLR (LLM/NLP) & ROC-AUC \metricup & 0.8258 & \textbf{0.8316} & 0.8292 \\
ICLR (RL) & ROC-AUC \metricup & 0.8053 & \textbf{0.8108} & 0.8056 \\
ICLR (Trustworthy AI) & ROC-AUC \metricup & 0.7659 & \textbf{0.7848} & 0.7781 \\
ICLR (Graphs) & ROC-AUC \metricup & 0.8036 & \textbf{0.8207} & 0.8070 \\
ELLIPSE & MAE \metricdown & 0.3334 & \textbf{0.3243} & 0.3254 \\
Persuasion & ROC-AUC \metricup & \textbf{0.8393} & 0.8385 & 0.8351 \\
MIMIC-MV& PR-AUC \metricup & 0.5218 & \textbf{0.5338} & 0.5234 \\
PAD-UFES-20 & ROC-AUC \metricup & \textbf{0.9734} & 0.9719 & 0.9720 \\
Danish Fungi &  ROC-AUC \metricup & \textbf{0.9310} & 0.9261 & 0.9190 \\
\bottomrule
\end{tabular}
\end{table}
Overall, the direct prediction score can further improve TableLoop, yielding higher held-out performance on six of the nine tasks.
Among the two integration strategies, feature concatenation outperforms score fusion on seven tasks, although often by a small margin.
This pattern favors incorporating the direct score at the feature level, where TabPFN can learn its relationship with the rubric features and metadata, over combining the final predictions post hoc.

\begingroup
\color{revisionblue}

\section{Examples of Rubric Evolution}
\label{app:rubric-examples}

We show selected rubric criteria from two tasks to illustrate how the features and scoring instructions change during evolution. For each task, we show the initial rubric, examples from intermediate iterations, and excerpts from the candidate with the highest rubric-only validation score. The scoring model assigns one numeric feature per criterion, and TabPFN learns how to use these features for prediction.

\subsection{MIMIC-MV}

The initial rubric asks for a single estimate of ventilation risk. By iteration 2, the rubric scores several sources of evidence separately, including respiratory distress and radiographic abnormalities. The validation-selected rubric gives more specific instructions about which findings to consider and how to weigh them. For example, it asks the scoring model to assess the physiological impact of atelectasis and to give more weight to imaging limitations when they accompany acute clinical concerns.

\paragraph{Initial rubric.}
\textbf{Ventilation risk score.} ``how likely this patient will require invasive mechanical ventilation (intubation) within 24 hours of this point in time''

\paragraph{Iteration 2 (selected criteria).}
\begin{itemize}[nosep]
    \item \textbf{Respiratory distress severity.} ``degree of acute respiratory compromise based on chief complaint and radiology findings, including presence of dyspnea, hypoxia, tachypnea, or signs of increased work of breathing''
    \item \textbf{Radiological abnormality burden.} ``extent and severity of acute intrathoracic abnormalities (e.g., consolidation, effusion, pneumothorax, edema, atelectasis) observed on imaging''
\end{itemize}

\paragraph{Validation-selected rubric (selected excerpts).}
\begin{itemize}[nosep]
    \item \textbf{Atelectasis physiological impact.} ``degree to which atelectasis appears to impair respiratory mechanics or gas exchange, based on radiographic signs such as bronchovascular crowding, obscured cardiac borders, volume loss, diaphragmatic elevation, or poor inspiration''
    \item \textbf{Imaging urgency as an instability proxy.} ``score higher when technical limitations co-occur with acute clinical concerns''
\end{itemize}

\subsection{ICLR Graphs}

This example uses paper text from PaperScorer-Text, as described in Appendix~\ref{app:paperscorer}. The initial rubric asks for overall paper quality. Iteration 1 introduces a criterion for handling graph structure, and iteration 5 names specific concepts such as graph inductive bias and expressiveness limitations associated with the 1-WL test. The validation-selected rubric goes further by specifying what evidence earns a high score. For example, it distinguishes a new expressiveness result supported by a proof from a reference to an established hierarchy.

\paragraph{Initial rubric.}
\textbf{Paper score.} ``overall quality of the paper''

\paragraph{Iteration 1 (selected criterion).}
\textbf{Handling of graph structure.} ``How effectively the method leverages or addresses graph-specific challenges (topology, symmetries, local vs.\ global information, sparsity, heterogeneity). Particularly relevant for GNN-based work.''

\paragraph{Iteration 5 (selected excerpts).}
\begin{itemize}[nosep]
    \item \textbf{Graph inductive bias.} ``Reward methods that explicitly model topology, symmetries, equivariance/invariance, subgraph patterns, hierarchical structure, or spectral properties.''
    \item \textbf{Architectural innovation.} ``Consider whether the method addresses fundamental expressiveness limitations (e.g., 1-WL test, over-smoothing, long-range dependencies).''
\end{itemize}

\paragraph{Validation-selected rubric (selected excerpt).}
\textbf{Theoretical expressiveness characterization.} ``High scores (85--100) require NEW proven results connecting the method to established hierarchies (WL, FWL, subgraph counting) with complete proofs, novel separation results, or constructive characterizations showing what graph properties the method can/cannot distinguish.'' The criterion also asks the scorer to assess whether the paper ``proves genuinely new theoretical results, or merely cites existing frameworks.''

\endgroup

\section{Naive versus Expert Rubric Initialization}
\label{sec:init-abla}
The outcome of prompt evolution may depend on the rubric used to initialize the search. We therefore compare two initializations for TableLoop with OpenEvolve. The \emph{naive initialization} contains only one broad, target-related dimension: overall paper quality for ICLR and overall writing proficiency for ELLIPSE. The \emph{expert initialization} starts from multiple task-specific dimensions. For ICLR, these dimensions are adapted from the official review criteria, including soundness, presentation, significance, originality, overall recommendation, and reviewer confidence. For ELLIPSE, they follow its human analytic scoring rubric and cover cohesion, syntax, vocabulary, phraseology, grammar, and conventions \citep{crossley2023ellipse}. We keep all other settings fixed and run both variants for 50 iterations. We restrict this comparison to ICLR and ELLIPSE because both have established, externally defined evaluation criteria: the ICLR review criteria and the ELLIPSE human scoring rubric. These provide a basis for expert initialization without requiring us to design a task-specific expert rubric, whose quality could otherwise confound the comparison.

\begin{table}[H]
\centering
\small
\caption{\textbf{Effect of rubric initialization on TableLoop with OpenEvolve.}
Both variants use the same fixed 50-iteration budget; ICLR results are averaged over three folds.}
\label{tab:init-ablation}
\begin{tabular}{@{}lccccc@{}}
\toprule
Initialization
& ICLR (LLM/NLP)
& ICLR (RL)
& ICLR (Trustworthy AI)
& ICLR (Graphs)
& ELLIPSE \\
\cmidrule(lr){2-5}
\cmidrule(l){6-6}
& \multicolumn{4}{c}{AUC \metricup}
& MAE \metricdown \\
\midrule
Naive
& \textbf{0.8258}
& \textbf{0.8053}
& \textbf{0.7659}
& \textbf{0.8036}
& \textbf{0.3334} \\
Expert
& 0.7848
& 0.7693
& 0.7379
& 0.7800
& 0.3349 \\
\bottomrule
\end{tabular}
\end{table}

As shown in Table \ref{tab:init-ablation}, naive initialization performs better on ELLIPSE and all four ICLR categories. This suggests that a detailed initial rubric is not necessarily a better starting point for evolution. A predefined decomposition may restrict the directions explored during optimization, whereas a single broad dimension leaves OpenEvolve more flexibility to discover and reorganize dimensions based on validation feedback. This supports our use of naive initialization in the main experiments.

\section{Direct Rubric Evolution versus Instruction Evolution}
\label{sec:instruction-evolution}
We test whether predictive feedback is more useful for revising a rubric directly or for revising instructions that generate a rubric. TableLoop takes the first approach: the text being evolved defines the features and how to score them. For comparison, we implement an \emph{instruction evolution} variant inspired by \citet{cosma2026feature}, in which OpenEvolve instead revises a prompt describing how a rubric should be constructed. A separate language-model call then uses that prompt to generate the rubric.

Both approaches use 50 OpenEvolve iterations and the same scoring and TabPFN evaluation pipeline. In the instruction-evolution variant, each rubric-generation call also receives 16 validation examples selected using the parent candidate's predictions: the four best-predicted and four worst-predicted examples from each class. The generated rubric is scored and evaluated, and its validation performance guides subsequent revisions to the generation instructions. Neither approach uses test examples or labels during optimization.

\begin{table}[H]
\centering
\small
\caption{\textbf{Direct rubric evolution outperforms instruction evolution on both tested tasks.} Both approaches use 50 OpenEvolve iterations and a three-rubric ensemble selected by rubric-only validation performance. Structured measurements are added when fitting the final TabPFN predictors. Higher values are better.}
\label{tab:instruction-ablation}
\begin{tabular}{lccc}
\toprule
Task & Metric & TableLoop & Instruction evolution \\
\midrule
Persuasion & ROC-AUC \metricup & \textbf{0.8393} & 0.8316 \\
MIMIC-MV & PR-AUC \metricup & \textbf{0.5218} & 0.4947 \\
\bottomrule
\end{tabular}
\end{table}

Direct rubric evolution achieves higher held-out performance on both tasks, improving ROC-AUC by $0.0077$ on Persuasion and PR-AUC by $0.0271$ on MIMIC-MV (Table \ref{tab:instruction-ablation}). These results support revising the scoring criteria and instructions directly. One possible explanation is that direct evolution lets predictive feedback guide changes to the exact rubric being evaluated, whereas instruction evolution must influence that rubric through an additional generation step. The comparison supports this design choice on the two tested tasks, but does not establish the cause of the performance difference.

\section{Reducing Evolution Cost through Training Subsampling}
\label{sec:train-subsampling}

Evaluating a candidate rubric requires applying the scoring model to both training and validation examples, making scoring the dominant cost of evolution. We test whether this cost can be reduced by subsampling the training set while retaining the full validation set for candidate evaluation, selection, and error feedback. We conduct this ablation on ELLIPSE using the same naive initialization and 50-iteration OpenEvolve budget as the main experiment. For the subsampled runs, a new training subset is drawn for each candidate evaluation, while validation feedback is sampled from pooled high- and low-error examples. Final top-three ensembles are evaluated using the full training and test sets.

\begin{table}[H]
\centering
\small
\caption{\textbf{Training subsampling preserves ELLIPSE performance.}
Held-out test MAE for OpenEvolve using different fractions of the training set during evolution. Validation always uses the full split. Lower values are better.}
\label{tab:train-subsampling}
\begin{tabular}{lcc}
\toprule
Training fraction & Rubric features & + structured measurements \\
\midrule
10\%  & 0.3463 & 0.3323 \\
30\%  & 0.3453 & 0.3348 \\
50\%  & 0.3517 & 0.3387 \\
100\% & 0.3450 & 0.3334 \\
\bottomrule
\end{tabular}
\end{table}

Across the tested fractions, training subsampling maintains competitive test performance while reducing the number of examples scored during evolution. The 10\% run even slightly improves MAE after structured measurements are added. Training subsampling therefore provides a practical way to reduce evolution cost, although aggressive subsampling may be less stable on datasets with fewer training examples.

\begin{table}[H]
\centering
\small
\caption{\textbf{Fixed and resampled training subsets give similar results.}
ELLIPSE test MAE with a 30\% training subset during evolution. Lower values are better.}
\label{tab:fixed-resampled-subsampling}
\begin{tabular}{lcc}
\toprule
Sampling strategy & Rubric features & + structured measurements \\
\midrule
Fixed subset     & \textbf{0.3430} & \textbf{0.3337} \\
Resampled subset & 0.3459 & 0.3344 \\
\bottomrule
\end{tabular}
\end{table}

Fixed and resampled training subsets produce similar results, suggesting that the choice between these two sampling strategies has little effect in this experiment.

\section{Feature Importance Feedback}

We test whether adding global mean-absolute SHAP importance for each rubric dimension to OpenEvolve validation feedback improves Persuasion.
The comparison uses the 50-iteration OpenEvolve \topthree{} with TabPFN.

\begin{table}[H]
\centering
\small
\caption{Persuasion feature-importance feedback (50-iteration OpenEvolve, TabPFN \topthree{}, ROC-AUC \metricup).}
\label{tab:fi-feedback}
\begin{tabular}{lccc}
\toprule
Evaluation & Control & + FI feedback & Difference \\
\midrule
Rubric features only & 0.8135 & 0.8100 & $-0.0035$ \\
+ metadata & 0.8393 & 0.8455 & $+0.0062$ \\
\bottomrule
\end{tabular}
\end{table}

The effect is small and changes sign once metadata is added, so we do not include SHAP importance in the default evolution feedback.

\section{LLM-Generated Confidence Features}

For MIMIC-MV, we additionally ask Qwen3-235B to report a confidence score, calibrate it on validation data, and add it to the rubric features.
OpenEvolve uses the 50-iteration \topthree{}; GEPA uses the \topthree{} from the first 30 iterations.

\begin{table}[H]
\centering
\small
\caption{MIMIC-MV confidence-feature experiment with TabPFN and the \topthree{} (OpenEvolve 50 iterations, GEPA 30 iterations).}
\label{tab:confidence}
\begin{tabular}{llccc}
\toprule
Evolution method & Added features & Shot & ROC-AUC & PR-AUC \\
\midrule
OpenEvolve & Confidence & 0 & 0.8043 & 0.3382 \\
& Confidence & 10 & 0.8069 & 0.3464 \\
& Metadata + confidence & 0 & 0.8427 & 0.5314 \\
& Metadata + confidence & 10 & 0.8464 & 0.5344 \\
& Metadata + confidence + direct score & 0 & 0.8472 & 0.5121 \\
& Metadata + confidence + direct score & 10 & 0.8472 & 0.5312 \\
\midrule
GEPA & Confidence & 0 & 0.8035 & 0.3648 \\
& Confidence & 10 & 0.8102 & 0.3816 \\
& Metadata + confidence & 0 & 0.8485 & 0.5485 \\
& Metadata + confidence & 10 & 0.8533 & 0.5590 \\
& Metadata + confidence + direct score & 0 & 0.8579 & 0.5712 \\
& Metadata + confidence + direct score & 10 & 0.8538 & 0.5597 \\
\bottomrule
\end{tabular}
\end{table}

Confidence yields mixed gains: they are modest without metadata, larger in some metadata settings, and not monotone in shot count when the direct prediction score is also included.
Because the pattern is unstable across evolution methods and feature packages, we do not include confidence in the default TableLoop representation.

\section{Experimental Setup Details}
\label{app:experimental-setup}

\subsection{Datasets}

\paragraph{ICLR papers.}
Using the topic labels provided by the ICLR dataset \citep{gonzalez2024learning}, we define four acceptance-prediction tasks: Graphs, Reinforcement Learning, Trustworthy AI, and LLM/NLP. Each task has its own evolved rubrics. Within each category, all ICLR 2025 papers form the training set, while ICLR 2026 papers are divided into validation and test sets under three stratified folds. The unstructured input consists of the title, abstract, and body text parsed with Docling \citep{auer2024docling}, truncated at 12,000 scoring-model tokens. Structured features comprise the numbers of authors, figures, and tables, and appendix length. The target is acceptance versus rejection. We compute test ROC-AUC separately for each fold and report the mean across the three folds.

\paragraph{ELLIPSE.}
ELLIPSE contains essays written by English-language learners, with human-assigned Overall proficiency scores from 1.0 to 5.0 in increments of 0.5 \citep{crossley2023ellipse}. We predict this score using the writing prompt and essay text alongside nine structured features describing length, lexical diversity, readability, paragraph structure, and prompt type. We partition the released training data into training and validation sets and retain the official test set for held-out evaluation. Performance is measured by mean absolute error (MAE).

\paragraph{Persuasion For Good.}
Persuasion For Good contains dialogues in which one participant attempts to persuade another to donate part of their task payment to Save the Children \citep{wang2019persuasion}. We predict whether the persuadee makes an actual donation using the full dialogue transcript and structured features describing dialogue statistics and participant demographics, personality, moral foundations, and values. All dialogues involving the same persuadee are assigned to the same split. Performance is measured by ROC-AUC.

\paragraph{MIMIC-MV.}
We construct this dataset from MIMIC-IV, MIMIC-IV-ED, and MIMIC-IV-Note \citep{johnson2023mimiciv,johnson2023mimicived,johnson2023mimicivnote}. The unstructured input contains the emergency-department chief complaint and radiology report. Structured features include demographics, triage measurements, the last recorded ED vital signs, and laboratory values. Taking the radiology-report time as the prediction time, we predict whether invasive ventilation begins within the following 24 hours. All positive ventilation events occur after the report time. To reduce textual leakage, we additionally exclude instances whose reports explicitly mention an endotracheal tube, intubation, tracheostomy, or mechanical ventilation. Because positive outcomes are uncommon, performance is measured by PR-AUC.

\paragraph{PAD-UFES-20.}
PAD-UFES-20 contains 2,298 smartphone clinical images of 1,891 lesions from 1,373 patients \citep{pacheco2020pad}. We predict cancer---basal cell carcinoma, squamous cell carcinoma, or melanoma---versus actinic keratosis, nevus, or seborrheic keratosis. Inputs comprise a lesion image and 21 structured variables describing demographics, lesion characteristics, symptoms, risk factors, and medical history. We use a deterministic 70/15/15 training/validation/test split, grouping instances that share a patient, lesion, or byte-identical image to prevent overlap across splits. For lesions with multiple images, we average prediction scores before computing lesion-level ROC-AUC.

\paragraph{Danish Fungi.}
We use field photographs and observation records from Danish Fungi 2020 \citep{picek2022danish}. The task is to identify one of five visually similar Russula species: \emph{Russula emetica}, \emph{Russula paludosa}, \emph{Russula rosea}, \emph{Russula nobilis}, and \emph{Russula velenovskyi}. The subset contains 1,135 images from 654 observations. Inputs comprise a photograph and three structured variables describing habitat, substrate, and observation month. We use a deterministic 70/15/15 training/validation/test split, stratified by species and assigning every image from the same observation to the same split. For observations with multiple images, we average prediction scores before computing observation-level macro ROC-AUC.

\subsection{TableLoop Configuration}

\paragraph{Models.}
For rubric evolution, we use Claude Sonnet 4.5 at temperature 0.7 on all tasks except MIMIC-MV. Claude Sonnet 4.5 was released before the ICLR 2026 acceptance decisions were announced, reducing the risk of exposure to those decisions during model training. For MIMIC-MV, we use locally hosted Qwen3-235B to keep clinical data within our local environment under the dataset's data-use restrictions. For scoring, we use Qwen3-32B on text tasks \citep{yang2025qwen3} and Qwen3-VL-32B-Instruct on image tasks, both at temperature 0.1. Each instance is scored three times, and responses are averaged separately for each rubric dimension. All main experiments use TabPFN with default estimator settings and random seed 42. No language-model weights are updated.

\paragraph{Shared evolution settings.}
All methods start from a naive rubric containing one dimension. Candidate evaluation and validation-based rubric selection use rubric scores alone. Structured measurements are added only after selection, when fitting the final tabular predictors. We reject candidates if fewer than $95\%$ of their combined training and validation scoring responses parse successfully. Missing rubric values are imputed using training-set medians.

\paragraph{OpenEvolve.}
We use a population size of 50, an archive size of 5, and three islands, with migration every two iterations. Feedback includes the parent's validation performance and the five highest-error validation instances, ranked by confident mistakes on classification tasks and by absolute residual on regression. Each run stops after 50 iterations.

\paragraph{GEPA.}
We use metric-aligned epsilon-greedy selection with $\epsilon=0.3$, a reflection minibatch of 10 validation instances, and no merge operator. Each run stops after 30 full-evaluation iterations, giving a scoring-model call budget comparable to 50 OpenEvolve iterations.

\paragraph{SH-merge.}
We begin with $C=64$ candidates and use $D=5$ evaluation rounds. The initial per-candidate sample budget is
\begin{equation}
B=\left\lceil
\frac{N_{\mathrm{train}}+N_{\mathrm{val}}}{2^{D-1}}
\right\rceil,
\end{equation}
where $N_{\mathrm{train}}$ and $N_{\mathrm{val}}$ denote the split sizes. With five rounds, the initial budget is one sixteenth of the combined training and validation set size, rounded up. The budget doubles between rounds, capped at the full dataset, while the candidate population halves through merging. The five rounds therefore evaluate 64, 32, 16, 8, and 4 candidates, respectively. The final round uses all training and validation instances in their respective roles, and the three best rubrics are selected from its four candidates.

We ran each experiment, varying the downstream tabular predictor and evolution method once, with a consistent random seed.

\subsection{Baselines}

\paragraph{Metadata-only.}
This baseline applies TabPFN to the structured features alone. It uses the same data splits and TabPFN configuration as TableLoop, isolating the predictive contribution of the rubric features.

\paragraph{Direct prediction.}
The direct-prediction baseline receives the task description, unstructured input, and structured features, then outputs a probability for classification or a scalar prediction for regression. We use Claude Sonnet 4.5 for ICLR, ELLIPSE, Persuasion For Good, PAD-UFES-20, and Danish Fungi. For MIMIC-MV, we use locally deployed Qwen3-235B to keep the clinical data within our local environment. All calls use zero-shot prompting at temperature 0. We fit an affine calibration map $\widetilde y=as+b$ to validation predictions and targets, then apply it unchanged to test predictions. Reported performance is computed from the calibrated test predictions.

\paragraph{FEST.}
FEST (Feature Engineering with Self-Evolving Trees) \citep{khurana2026fest} discovers interpretable features from contrasting positive and negative examples through two complementary streams. Its semantic features are natural-language predicates whose presence is assessed by a language model, while its deterministic features are implemented as executable code. At each iteration, FEST consolidates semantically redundant candidates, adds the resulting features to a shared bank, trains a decision tree on their encodings, and removes features with consistently low importance. We reimplement this discovery-mode procedure without expert-provided feature seeds, following the algorithmic structure and prompts described in the original paper. The original implementation uses GPT-4o-mini for its language-model stages and Qwen3-Embedding-4B for semantic consolidation; we use Claude Sonnet 4.5 for feature generation and consolidation, Qwen3-32B for semantic-feature encoding, Qwen3-VL-32B-Instruct for image inputs, and the same Qwen3-Embedding-4B model. For MIMIC-MV, all language-model stages use locally deployed Qwen models. To match the TableLoop evaluation protocol, we concatenate the selected FEST features with the same structured features and fit the same TabPFN predictor, rather than the classifier suite used in the original paper. We evaluate FEST on the seven binary classification tasks using the same training, validation, and test splits as TableLoop.
\begingroup
\color{revisionblue}
\subsection{Additional ICLR Results with PaperScorer-Text}
\label{app:paperscorer}

We also reran our ICLR experiments using the newly released PaperScorer-Text dataset \citep{konan2026paperscorer} as an alternative source of paper text. The dataset provides anonymized Markdown extracted with MinerU from the latest publicly available pre-camera-ready OpenReview PDFs. We match papers from our original ICLR evaluation to PaperScorer-Text by OpenReview submission ID and use the released \texttt{content} field as the text source. The retained papers keep their original labels, metadata, and split assignments in each fold.

We reran TableLoop with
OpenEvolve, zero-shot direct prediction, and metadata-only TabPFN
on ICLR (Graphs) and ICLR (RL). We use the same method configurations
and evaluation protocol as in the main experiments. TableLoop continues
to outperform both baselines on both tasks
(Table~\ref{tab:paperscorer-results}).

\begin{table}[H]
\centering
\color{revisionblue}
\caption{\textbf{Additional ICLR results using PaperScorer-Text.}
Test ROC-AUC~\metricup{} is averaged across three folds.}
\label{tab:paperscorer-results}
\begin{tabular}{lcc}
\toprule
Method & ICLR (Graphs) & ICLR (RL) \\
\midrule
TableLoop (OpenEvolve)           & \textbf{0.6946} & \textbf{0.7014} \\
LLM direct prediction baseline & 0.6030         & 0.6076 \\
Metadata-only baseline             & 0.6204         & 0.5808 \\
\bottomrule
\end{tabular}
\end{table}

\par
\endgroup

\begingroup
\color{revisionblue}

\section{Fine-Tuned Baselines}

\subsection{AutoMM}
\label{app:automm}

We compare TableLoop with AutoGluon-Multimodal (AutoMM)
\citep{tang2024automm}, a supervised multimodal AutoML framework
that fine-tunes pretrained encoders and fuses text or image
representations with tabular features for prediction. AutoMM and TableLoop use the same source inputs, metadata, data splits,
and evaluation metrics. For the two ICLR tasks, the source text comes
from PaperScorer-Text, as described in Appendix~\ref{app:paperscorer}.

All AutoMM runs use version 1.6.1, the \texttt{best\_quality} preset,
and seed 42. Data augmentation is enabled at the default scale of 0.1
for both text and images. Models are fit on the
training split, with the validation split used for early stopping
and model selection. The default text configuration uses
DeBERTa-v3-base \citep{he2021debertav3} with CLS pooling and a 512-token limit. We also run a long-context variant of the \texttt{best\_quality}
preset, replacing DeBERTa-v3-base with ModernBERT-base \citep{warner-etal-2025-smarter}, switching
to mean pooling, and increasing the token limit to 8,192 tokens. The image tasks use the preset's Swin-Large
encoder \citep{Liu_2021_ICCV} at $224 \times 224$ resolution.

TableLoop achieves higher ROC-AUC than either AutoMM configuration
on Persuasion and both ICLR tasks (Table~\ref{tab:automm-comparison}).
The differences are largest on Persuasion and ICLR (RL), where
TableLoop exceeds the stronger AutoMM configuration by 0.1279 and
0.1152 ROC-AUC, respectively. The two approaches obtain similar
scores on MIMIC-MV and PAD-UFES-20, while AutoMM performs better on
ELLIPSE and Danish Fungi.
\begin{table}[H]
\centering
\small
\color{revisionblue}
\caption{\textbf{Comparison with supervised AutoMM.} For text tasks, Default uses DeBERTa-v3-base. Long-context replaces it with ModernBERT-base with a higher token limit. ICLR results are
mean test ROC-AUC across the three folds.}
\label{tab:automm-comparison}
\begin{tabular}{llccc}
\toprule
& & \multicolumn{2}{c}{AutoMM} & \\
\cmidrule(lr){3-4}
Task & Metric & Default & Long-context & TableLoop (OpenEvolve) \\
\midrule
Persuasion   & ROC-AUC $\uparrow$ & 0.7037 & 0.7114    & 0.8393 \\
ELLIPSE      & MAE $\downarrow$   & 0.3270 & 0.3207    & 0.3334 \\
MIMIC-MV     & PR-AUC $\uparrow$  & 0.5159 & 0.5238    & 0.5218 \\
ICLR (RL)     & ROC-AUC $\uparrow$ & 0.5684 & 0.5862 & 0.7014 \\
ICLR (Graphs) & ROC-AUC $\uparrow$ & 0.6509 & 0.6613 & 0.6946 \\
PAD-UFES-20  & ROC-AUC $\uparrow$ & 0.9750 & \textemdash & 0.9734 \\
Danish Fungi & ROC-AUC $\uparrow$ & 0.9600 & \textemdash & 0.9310 \\
\bottomrule
\end{tabular}
\end{table}

Across the comparisons with AutoMM and frozen embeddings
(Section~\ref{sec:embedding-comparison}), TableLoop's gains are most
pronounced on Persuasion and ICLR acceptance prediction. This pattern
suggests that rubric-based scoring may be particularly useful for
tasks that require reasoning over text, where the rubric guides the
scorer to assess evidence relevant to the prediction target. Adapting the rubric also keeps the scoring criteria explicit and
interpretable while allowing TableLoop to reuse a frozen scorer
across tasks without maintaining a separate fine-tuned encoder
checkpoint for each task.

\subsection{LoRA Fine-Tuning}
\label{app:finetuned-llm}
We evaluated whether TableLoop could outperform direct prediction with LLMs fine-tuned for the specific prediction task we evaluated. We fine-tuned Qwen3-32B with LoRA \citep{hu2021lora} to directly predict whether the
persuadee makes a donation in Persuasion For Good. The model receives
the same dialogues and ten metadata features as TableLoop and uses
the same training, validation, and test splits. It is trained to
output \texttt{1} for a donation and \texttt{0} otherwise, with
cross-entropy loss computed only on the answer token. At evaluation,
we apply a softmax to the logits for \texttt{0} and \texttt{1} at
the answer position and use the resulting probability of \texttt{1}
to compute ROC-AUC.

We freeze the base model weights and apply LoRA adapters to the
query, key, value, and output projections with rank~16, $\alpha=32$,
and dropout~0.05, yielding 39.8 million trainable parameters.
We use AdamW with a learning rate of $10^{-4}$, a batch size of one,
16 gradient accumulation steps, and gradient clipping at 1.0.
Training uses seed~42 and runs for up to 15 epochs, with early
stopping after three consecutive epochs without improvement in
validation ROC-AUC. The checkpoint with the highest validation
ROC-AUC is evaluated once on the test set.

On Persuasion For Good, the LoRA-fine-tuned Qwen3-32B baseline
achieves a test ROC-AUC of 0.7930. TableLoop with OpenEvolve achieves
0.8393, approximately 0.0463 higher than the fine-tuned baseline,
without updating the scoring model's weights.

\section{Does TableLoop's Performance Benefit come from Optimization or Tabularization?}

To test whether TableLoop's performance improvements over direct LLM prediction come solely from tabularization versus from the optimization of our prediction pipeline, we conducted two experiments to isolate the effects. First, we generated rubrics in a zero-shot manner with the same dimension as our final rubric (i.e., same tabularization, no evolutionary optimization). Second, we used the same evolutionary optimization methods from TableLoop to optimize an LLM prompt for direct prediction (same optimization, no tabularization).

\subsection{Rubric Generation with Fixed Dimensionality}
\label{app:fixed-dimension-rubrics}

We evaluate the performance of zero-shot rubric generation and a
one-round variant of TableLoop when both are given the dimensionalities
of the top-3 rubrics from the full
50-round OpenEvolve run. For each task, we record the
dimensionalities of the top-3 OpenEvolve rubrics as $d_1$, $d_2$, and
$d_3$. Each baseline generates three rubrics in separate calls, with
the $i$-th call required to produce a rubric with exactly $d_i$
scoring dimensions. Since finding an appropriate
dimensionality may itself be part of what evolution contributes,
this setup gives both baselines the dimensionality information
obtained through the search.

Zero-shot generates a rubric from the task description and
rubric format, without an initial rubric or validation
feedback. TableLoop (one round) rewrites the same initial
rubric used by full TableLoop once, using its validation score and
the five highest-error validation examples as feedback. Full
TableLoop can change the rubric dimensionality during search.
Across all settings, we use the same scoring models and fit one
TabPFN predictor per rubric using its scores and the metadata
from the main experiments. We average the three predictions for
each example before computing the test metric. Both
ICLR tasks use the PaperScorer-Text setup described in
Appendix~\ref{app:paperscorer}.
\begin{table}[H]
\centering
\small
\color{revisionblue}
\caption{\textbf{Comparison with fixed-dimensional rubric baselines.}
All methods use metadata. ICLR results are averaged across three test folds.}
\label{tab:fixed-dimension-rubrics}
\begin{tabular}{llccc}
\toprule
Task & Metric & Zero-shot & TableLoop (one round) & TableLoop (full) \\
\midrule
Persuasion    & ROC-AUC $\uparrow$ & 0.7988 & 0.8264 & \textbf{0.8393} \\
ELLIPSE       & MAE $\downarrow$  & 0.3351 & 0.3384 & \textbf{0.3334} \\
MIMIC-MV      & PR-AUC $\uparrow$ & 0.4870 & 0.5190 & \textbf{0.5218} \\
ICLR (RL)     & ROC-AUC $\uparrow$ & 0.6671 & 0.6747 & \textbf{0.7014} \\
ICLR (Graphs) & ROC-AUC $\uparrow$ & 0.6419 & 0.6345 & \textbf{0.6946} \\
PAD-UFES-20   & ROC-AUC $\uparrow$ & 0.9708 & 0.9714 & \textbf{0.9734} \\
Danish Fungi  & ROC-AUC $\uparrow$ & 0.9298 & \textbf{0.9310} & \textbf{0.9310} \\
\bottomrule
\end{tabular}
\end{table}

Full TableLoop outperforms both baselines on six of seven tasks
and matches the one-round variant on Danish Fungi
(Table~\ref{tab:fixed-dimension-rubrics}). Its gains over the
stronger baseline on each task are largest on Persuasion and the
two ICLR acceptance-prediction tasks. This pattern suggests that iterative rubric refinement may be
particularly useful when prediction requires reasoning about
evidence across a text.

\subsection{Optimized Direct Prediction}
\label{app:optimized-direct-prediction}

To test whether TableLoop's advantage persists against a direct
predictor that also undergoes iterative task adaptation, we optimize
direct prediction prompts with GEPA \citep{agrawal2025gepa} on
Persuasion, ELLIPSE, and PAD-UFES-20. For this comparison, we use
the GEPA variant of TableLoop so that both approaches use the same
search algorithm.

The comparison uses the same training, validation, and test splits
and the same metadata features for each task. Claude Sonnet~4.5 proposes revisions in both searches. During search,
the same Qwen model on each task produces direct predictions for the
baseline and rubric scores for TableLoop. Each GEPA reflection step uses a
minibatch of 10 labeled validation examples, and each search has a
budget of 30 full-validation evaluations. Both
approaches select three candidates by validation performance and
average their predictions per example.

The direct predictor receives the target example's input and
metadata during search and at test time. The 0-shot variant uses no
demonstrations, while the 5-shot variant retrieves five training
examples by embedding similarity and includes each example's input,
metadata, and label. At test time, we use Claude Sonnet~4.5 to make predictions with
each of the three selected direct prediction prompts and average
the predictions for each test example. TableLoop likewise averages
predictions from three TabPFN predictors, each fit on the scores
from one selected rubric combined with metadata.

\begin{table}[H]
\centering
\color{revisionblue}
\caption{\textbf{GEPA-optimized direct prediction and TableLoop.}
Direct predictions are affinely calibrated on the
validation set before computing test metric.}
\label{tab:optimized-direct-prediction}
{\small
\begin{tabular}{llccc}
\toprule
& & \multicolumn{2}{c}{Direct prediction (GEPA)} & \\
\cmidrule(lr){3-4}
Task & Metric & 0-shot & 5-shot & TableLoop (GEPA) \\
\midrule
Persuasion  & ROC-AUC $\uparrow$ & 0.7998 & 0.8025 & \textbf{0.8275} \\
ELLIPSE     & MAE $\downarrow$   & 0.3797 & 0.3619 & \textbf{0.3358} \\
PAD-UFES-20 & ROC-AUC $\uparrow$ & 0.8734 & 0.9337 & \textbf{0.9733} \\
\bottomrule
\end{tabular}
\par}
\end{table}

TableLoop outperforms both optimized direct predictors on all three
tasks (Table~\ref{tab:optimized-direct-prediction}). Relative to the
stronger direct baseline on each task, TableLoop improves ROC-AUC
by 0.0250 on Persuasion and 0.0396 on PAD-UFES-20, and reduces MAE
by 0.0261 on ELLIPSE. These results show that TableLoop's advantage
persists when direct prediction receives the same data splits and
metadata, uses the same GEPA evaluation budget and top-three
ensembling, and has access to labeled training demonstrations in
the 5-shot setting.

\section{Runtime and Computational Cost}
\label{app:computational-cost}

TableLoop has two computational stages: offline rubric search and
online prediction with the selected rubrics. We first benchmark the scoring models and baselines by measuring
per-example GPU time with one input at a time and throughput with
concurrent requests or batched inputs. We then break down the costs of offline search and online
prediction and report full-split timings for one search iteration and
one held-out evaluation with a selected rubric. Throughout, an operation
that uses $g$ GPUs for $t$ seconds consumes $gt$ GPU-seconds.

\subsection{Per-Example GPU Time and Throughput}

For an autoregressive scoring request with rubric $q$ and input $x$,
we decompose its wall-clock latency as
\begin{equation}
t_{\mathrm{score}}(q,x)
=
t_{\mathrm{TTFT}}(q,x)
+
t_{\mathrm{decode}}(q,x).
\end{equation}
Here $t_{\mathrm{TTFT}}$ is the time from sending the request to
receiving the first output token, including prompt prefill.
The decoding time $t_{\mathrm{decode}}$ is the time after the first
token, during which the model generates the remaining output tokens
autoregressively.

Table~\ref{tab:gpu-seconds} reports mean GPU-seconds per example,
with one example processed at a time. We group the test examples into quartiles
by whitespace-delimited token count for text and by file size for
images, then sample 25 examples from each quartile using seed~42. All models are then timed on this
same 100 test examples per dataset, following an untimed warm-up
on five training examples. 
For ICLR, the 100 examples are sampled from a single pool of
PaperScorer-Text test examples across all four ICLR
categories.

The embedding models encode the text or images without generating output tokens. The 235B models make a zero-shot
prediction from the input and metadata.
The 8B and 32B Qwen scorers apply the highest-ranked OpenEvolve rubric and generate one JSON response containing all
rubric scores for each example. We also benchmark System~1 of AutoTrust's JEV-9B
\citep{autotrust_jev9b_2026} as an alternative rubric scorer.
It applies the same rubrics but makes a separate one-token decision
for each dimension. For each example, we score the dimensions in
successive batches of up to eight concurrent requests and sum the
elapsed times across batches. The JEV column reports the mean of
these per-example totals across the 100 test examples.

\begin{table}[H]
\centering
\setlength{\tabcolsep}{2.4pt}
\color{revisionblue}
\caption{\textbf{Per-example GPU-seconds with one input at a time.}
All models use one L40S except 235B direct prediction, which uses
eight L40S GPUs with tensor and expert parallelism. GPU-seconds are
computed by multiplying elapsed time by the number of GPUs used.
TTFT measures time to first token, and Dec measures the time to
autoregressively generate the remaining tokens. JEV S1 reports the
mean total time to score all rubric dimensions for one example.
Image rows use the corresponding vision-language models.}
\label{tab:gpu-seconds}
{\small
\begin{tabular}{@{}lccccccccccc@{}}
\toprule
& Emb.
& \multicolumn{3}{c}{235B direct}
& \multicolumn{3}{c}{Qwen3-8B}
& \multicolumn{3}{c}{Qwen3-32B}
& JEV S1 \\
\cmidrule(lr){2-2}
\cmidrule(lr){3-5}
\cmidrule(lr){6-8}
\cmidrule(lr){9-11}
\cmidrule(lr){12-12}
Task & Total
     & Total & TTFT & Dec
     & Total & TTFT & Dec
     & Total & TTFT & Dec
     & Total \\
\midrule
Persuasion   & 0.062 &  4.08 &  2.14 &  1.94 &  5.72 & 0.10 &  5.62 & 16.87 & 0.27 & 16.59 & 0.92 \\
ELLIPSE      & 0.049 &  3.16 &  1.73 &  1.43 &  2.85 & 0.08 &  2.77 &  7.95 & 0.24 &  7.71 & 0.63 \\
MIMIC-MV     & 0.035 &  3.26 &  1.47 &  1.79 &  2.13 & 0.06 &  2.07 &  6.27 & 0.16 &  6.11 & 0.31 \\
ICLR         & 1.14  & 27.84 & 26.04 &  1.80 &  5.47 & 1.19 &  4.28 & 15.33 & 4.09 & 11.24 & 1.97 \\
PAD-UFES-20  & 0.200 &  5.00 &  3.08 &  1.92 &  3.83 & 0.22 &  3.61 &  8.48 & 0.45 &  8.03 & 0.99 \\
Danish Fungi & 0.087 & 12.53 &  1.52 & 11.01 & 12.06 & 0.25 & 11.82 & 29.52 & 0.70 & 28.82 & 1.98 \\
\bottomrule
\end{tabular}
\par}
\end{table}

We separately measure feature-extraction throughput on one L40S.
For each task and model, we sweep $\{1,2,4,8,16,32\}$ on 100
length-stratified training examples. These values denote concurrent
requests for Qwen scorers and batch sizes for embedding models.
We select the highest-throughput setting, breaking ties in favor
of the smaller value. We then use that setting to process the
same 100 test examples as in Table~\ref{tab:gpu-seconds} and report
the number of examples processed per second of wall-clock time.

\begin{table}[H]
\centering
\small
\setlength{\tabcolsep}{4pt}
\color{revisionblue}
\caption{\textbf{Feature-extraction throughput on one L40S.}
For Qwen rubric scorers, the selected setting is the number of
concurrent requests. For embedding models, it is the batch size.
Throughput is measured on 100 test examples.}
\label{tab:cost-concurrency-sweep}
\begin{tabular}{@{}llrr@{}}
\toprule
Task & Model & \shortstack{Concurrency\\or batch size}
     & \shortstack{Throughput\\(examples/s)} \\
\midrule
Persuasion   & Qwen3-Embedding-8B    & 32 & 19.601 \\
             & Qwen3-8B              & 32 &  1.710 \\
             & Qwen3-32B             & 32 &  0.625 \\
\midrule
ELLIPSE      & Qwen3-Embedding-8B    & 32 & 26.397 \\
             & Qwen3-8B              & 32 &  4.328 \\
             & Qwen3-32B             & 32 &  1.699 \\
\midrule
MIMIC-MV     & Qwen3-Embedding-8B    & 32 & 57.578 \\
             & Qwen3-8B              & 32 &  5.576 \\
             & Qwen3-32B             & 32 &  2.120 \\
\midrule
ICLR         & Qwen3-Embedding-8B    &  1 &  0.844 \\
             & Qwen3-8B              & 32 &  0.483 \\
             & Qwen3-32B             & 16 &  0.140 \\
\midrule
PAD-UFES-20  & Qwen3-VL-Embedding-8B &  2 &  5.177 \\
             & Qwen3-VL-8B           & 32 &  3.232 \\
             & Qwen3-VL-32B          & 32 &  1.404 \\
\midrule
Danish Fungi & Qwen3-VL-Embedding-8B & 16 & 59.674 \\
             & Qwen3-VL-8B           & 32 &  0.972 \\
             & Qwen3-VL-32B          & 32 &  0.355 \\
\bottomrule
\end{tabular}
\end{table}

Embedding extraction achieves higher throughput than Qwen rubric
scoring (Table~\ref{tab:cost-concurrency-sweep}). In the per-example
benchmark, autoregressive decoding accounts for most of the GPU time
for the Qwen rubric scorers. JEV System~1 (9B) scores the same rubrics with a roughly $5\times$
speedup over Qwen3-8B in per-example GPU time
(Table~\ref{tab:gpu-seconds}), suggesting that single-token decisions
could reduce the cost of rubric scoring.

\subsection{Offline Search and Online Prediction}

Table~\ref{tab:gpu-hours-one-rubric} reports GPU-hours on one L40S
for one offline candidate
evaluation and one held-out evaluation with a selected rubric. The offline measurement includes training- and validation-set scoring,
TabPFN fitting, and validation prediction. The held-out measurement
includes test-set scoring with the selected rubric, TabPFN fitting,
and test prediction.

Both measurements use three scoring generations per example, with
Qwen3-32B-FP8 for text and Qwen3-VL-32B-Instruct for images. For each dataset, we use the concurrency that achieved the highest
throughput for the corresponding 32B scorer in the preceding sweep
(Table~\ref{tab:cost-concurrency-sweep}). For both ICLR tasks, we
report timings on the PaperScorer-Text setup.

Offline search accounts for the repeated cost of evaluating new
rubrics on the training and validation data. In return, the search
learns explicit, task-specific criteria that turn semantic judgments
into interpretable and reusable features. Several strategies can reduce
this cost, including subsampling the training set at each iteration,
using one or two scoring generations per example instead of three,
and applying an early-stopping criterion. For example, on ELLIPSE, training-set
subsampling substantially reduces the number of scoring calls per
iteration while maintaining comparable predictive performance
(Appendix~\ref{sec:train-subsampling}).

\begin{table}[H]
\centering
\setlength{\tabcolsep}{4pt}
\color{revisionblue}
\caption{\textbf{GPU-hours for one offline candidate evaluation and
one held-out evaluation on one L40S.}
Train, Val., and Test denote rubric scoring on the corresponding
splits. The held-out evaluation uses one selected rubric.}
\label{tab:gpu-hours-one-rubric}
{\small
\begin{tabular}{@{}lccccccc@{}}
\toprule
& \multicolumn{4}{c}{Offline search (one iteration)}
& \multicolumn{3}{c}{Held-out evaluation (one rubric)} \\
\cmidrule(lr){2-5}
\cmidrule(lr){6-8}
Task & Train & Val. & TabPFN & Total
     & Test & TabPFN & Total \\
\midrule
Persuasion    & 0.9595 & 0.2055 & 0.0046 & 1.1696 & 0.1900 & 0.0372 & 0.2272 \\
ELLIPSE       & 1.3959 & 0.7010 & 0.0080 & 2.1049 & 1.1163 & 0.0055 & 1.1218 \\
MIMIC-MV      & 0.9518 & 0.2143 & 0.0052 & 1.1713 & 0.1581 & 0.0373 & 0.1954 \\
ICLR (RL)     & 2.6136 & 3.1140 & 0.0436 & 5.7712 & 1.5015 & 0.0409 & 1.5424 \\
ICLR (Graphs) & 1.8931 & 1.2225 & 0.0482 & 3.1637 & 0.5955 & 0.0408 & 0.6363 \\
PAD-UFES-20   & 0.9675 & 0.2103 & 0.0102 & 1.1880 & 0.1838 & 0.0046 & 0.1884 \\
Danish Fungi  & 1.6263 & 0.3581 & 0.0038 & 1.9881 & 0.4112 & 0.0022 & 0.4134 \\
\bottomrule
\end{tabular}
\par}
\end{table}
Once the rubrics are selected, online prediction requires only
scoring each new input with the three selected rubrics and averaging
the predictions from the fitted TabPFN models. Neither rubric search
nor training-set scoring is repeated, leading to lower computational
costs at deployment.

Furthermore, rubric scoring can be parallelized to shorten wall-clock
time during both offline search and online prediction. A faster scorer
could also reduce GPU time per example. In our per-example benchmark
(Table~\ref{tab:gpu-seconds}), autoregressive decoding accounts for most
of the Qwen rubric scorers' GPU time. JEV System~1 scores the same
rubrics using less GPU time by making one-token decisions for
individual dimensions instead of generating a multi-token JSON
response, making it a promising candidate for
reducing rubric scoring costs.

\endgroup
\end{document}

%% file: math_commands.tex
\usepackage{amsmath,amsfonts,bm}

\def\eqref#1{equation~\ref{#1}}

\def\1{\bm{1}}

\DeclareMathAlphabet{\mathsfit}{\encodingdefault}{\sfdefault}{m}{sl}
\SetMathAlphabet{\mathsfit}{bold}{\encodingdefault}{\sfdefault}{bx}{n}

